\documentclass{article}

\usepackage{PRIMEarxiv}

\usepackage[utf8]{inputenc} % allow utf-8 input
\usepackage[T1]{fontenc}    % use 8-bit T1 fonts
\usepackage{hyperref}       % hyperlinks
\usepackage{url}            % simple URL typesetting
\usepackage{booktabs}       % professional-quality tables
\usepackage{amsfonts}       % blackboard math symbols
\usepackage{nicefrac}       % compact symbols for 1/2, etc.
\usepackage{microtype}      % microtypography
\usepackage{lipsum}
\usepackage{fancyhdr}       % header
\usepackage{graphicx}       % graphics
\graphicspath{{media/}}     % organize your images and other figures under media/ folder
\usepackage{xcolor}         % colors
\usepackage{amsmath, amssymb}
\usepackage{algorithm}
\usepackage{algpseudocode}
\usepackage{multirow}
\usepackage{wrapfig}
\usepackage{xcolor}
\usepackage{arydshln}

\usepackage{natbib}

\usepackage[most]{tcolorbox}
\newtcolorbox{eqbox}{
  colback=orange!10, colframe=orange!50!black,
  boxrule=0.3pt, arc=1pt,
  left=4pt,right=4pt,top=0pt,bottom=0pt,
  before skip=1pt, after skip=1.2pt % <-- no extra vertical space
}

\lstdefinestyle{py}{
  language=Python,
  basicstyle=\ttfamily\small,
  numbers=left,
  numbersep=8pt,
  frame=single,
  breaklines=true,
  showstringspaces=false,
  tabsize=4,
}
\title{Stochastic Layer-wise Learning: Scalable and Efficient Alternative to Backpropagation 
}

\author{
  \textbf{Bojian Yin} \\
  Department of Electrical Engineering \\
  Eindhoven University of Technology \\
  The Netherlands \\
  \texttt{b.yin@tue.nl}
  \and
  \textbf{Federico Corradi} \\
  Department of Electrical Engineering \\
  Eindhoven University of Technology \\
  The Netherlands \\
  \texttt{f.corradi@tue.nl}
}

\begin{document}
\maketitle

\begin{abstract}
Backpropagation underpins modern deep learning, yet its reliance on global gradient synchronization limits scalability and incurs high memory costs. In contrast, fully local learning rules are more efficient but often struggle to maintain the cross-layer coordination needed for coherent global learning. Building on this tension, we introduce Stochastic Layer-wise Learning (SLL), a layer-wise training algorithm that decomposes the global objective into coordinated layer-local updates while preserving global representational coherence.  The method is ELBO-inspired under a Markov assumption on the network, where the network-level objective decomposes into layer-wise terms and each layer optimizes a local objective via a deterministic encoder. The intractable KL in ELBO is replaced by a Bhattacharyya surrogate computed on auxiliary categorical posteriors obtained via fixed geometry-preserving random projections, with optional multiplicative dropout providing stochastic regularization. SLL optimizes locally, aligns globally, thereby eliminating cross-layer backpropagation. Experiments on MLPs, CNNs, and Vision Transformers from MNIST to ImageNet show that the approach surpasses recent local methods and matches global BP performance while memory usage invariant with depth. The results demonstrate a practical and principled path to modular and scalable local learning that couples purely local computation with globally coherent representations.
\end{abstract}

% keywords can be removed

\section{Introduction}

The success of deep learning across a wide range of domains has been substantially driven by backpropagation (BP), a foundational learning algorithm enabling hierarchical representation learning through end-to-end gradient-based optimization~\cite{rumelhart1986learning, lecun2015deep}.  Despite its algorithmic clarity and practical effectiveness, BP requires the exact storage of intermediate activations and subsequent gradient computation across all layers. This mechanism facilitates global credit assignment~\cite{lillicrap2020backpropagation}; it also introduces a well-known bottleneck called \textit{update-locking}~\cite{jaderberg2017decoupled,griewank2008evaluating}, where the weight update of a given layer must wait until both the forward pass through the entire network and the backward pass through deeper layers are complete. Consequently, this global dependency limits asynchronous updating, and imposes substantial memory and computational overhead, ultimately reducing training efficiency and scalability, especially in resource-constrained devices~\cite{luo2024efficient,belilovsky2019greedy,bengio2006greedy}.

%Beyond efficiency constraints, BP is commonly regarded as biologically implausible, motivating the exploration to identify local learning rules that could achieve credit assignment in real neural systems ~\cite{lillicrap2020backpropagation,scellier2017equilibrium,guerguiev2017towards}. Recent advances in neuroscience propose that feedback connections in the brain can induce differences in local neural activity that approximate global error signals~\cite{whittington2019theories,guerguiev2017towards}, offering a bio-plausible pathway for distributed learning. These insights raise the possibility of uniting local synaptic plasticity with deep learning capabilities traditionally attributed to BP~\cite{lillicrap2020backpropagation,sacramento2018dendritic}. However, such mechanisms do not fully resolve the tension between local updates and global learning. Moreover, they lack a solid theoretical framework to support this integration, highlighting the need for principled alternatives to BP.

BP is often seen as biologically implausible and this drives efforts to discover local learning rules for credit assignment in inspired by real neural systems~\cite{lillicrap2020backpropagation, scellier2017equilibrium, guerguiev2017towards}. At the same time, neuroscience suggests that feedback connections may approximate global errors via local activity differences~\cite{guerguiev2017towards,whittington2019theories}, hinting at a bioplausible path to deep learning~\cite{lillicrap2020backpropagation,sacramento2018dendritic}. Yet, these approaches struggle to reconcile local updates with global learning and lack a unifying theoretical framework.

Given this context, a central research question emerges:
%\begin{quote}
\textit{``Can we design a theoretical framework capable of decomposing deep neural network training into local (layer-wise) optimizations while retaining the benefits of hierarchical representation learning?''}
%we\end{quote}
This question captures a fundamental conflict: while local learning encourages architectural scalability and computational parallelization, effective deep learning relies on non-linear coordination across the entire network. This disconnect often results in misaligned learning signal and suboptimal performance, challenging the network scalability of locally trained models~\cite{yang2024towards}.

To address this question, we ground local learning in how information propagates through deep networks: as signals traverse layers, raw inputs are progressively transformed into increasingly disentangled and class-separable representations~\cite{he2023law, razdaibiedina2022representation, telgarsky2016benefits}. This refinement suggests that intermediate layers perform latent inference, selectively preserving task-relevant signals while suppressing redundancy~\cite{shwartz2017opening}. In this paper, we make this intuition precise by exhibiting a network-level ELBO that decomposes into layer-wise terms under a Markov assumption of network architecture, thereby furnishing principled local objectives while retaining an explicit link to the global goal. Building on this decomposition, we introduce \textbf{Stochastic Layer-wise Learning (SLL)}, a local learning framework in which each layer produces auxiliary categorical posteriors via fixed stochastic random projections, and the intractable layer-wise KL in the ELBO is replaced by a Bhattacharyya surrogate~\cite{bhattacharyya1943} computed on these induced posteriors, yielding an ELBO-inspired and numerically stable update. Here, the projections preserve minibatch geometry with high probability by the Johnson--Lindenstrauss (JL) lemma ~\cite{johnson1984extensions, razdaibiedina2022representation}, which justifies computing divergences in the compressed space; we further apply multiplicative dropout to the fixed projection, which provides stochastic regularization consistent with the dropout-as-variational-inference interpretation~\cite{gal2016dropout}; we do not learn mask parameters and do not claim a variational bound over masks, and the overall objective remains ELBO-inspired at the layer level. SLL thus reconciles local optimization with hierarchical coordination, mitigating over-compression associated with direct KL minimization, maintaining global representational coherence, and enabling scalable, parallel training without full backpropagation.

 \begin{figure*}[t]
	\centering
	\includegraphics[width=.95\textwidth]{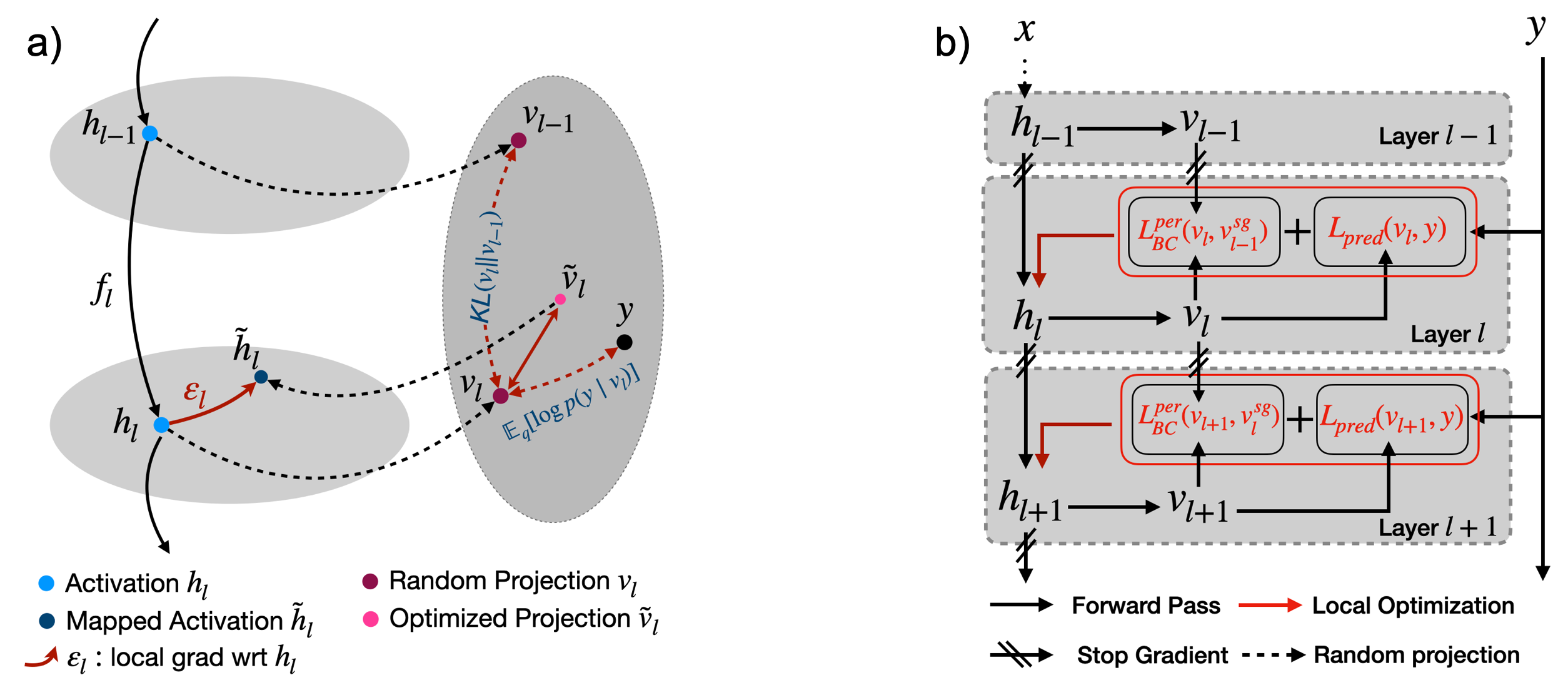}
    \caption{\textbf{Overview of Stochastic Layer-wise Learning (SLL).}
    (a) SLL treats each hidden activation $h_l$ as a latent variable and projects it to $v_l$ via a random matrix. The local ELBO comprises a log-likelihood term and a KL surrogate that promotes inter-layer consistency. Optimizing this loss yields the improved projection $\tilde v_l$ and its corresponding activation $\tilde h_l$.
    (b) SLL optimizes each layer independently using a prediction loss $\mathcal{L}_{\text{pred}}(v_l, y)$ from the log-likelihood and a feature alignment loss $\mathcal{L}_{\text{BC}}^{per}(v_l, v_{l-1}^{\text{sg}})$ approximating the KL term. 
    Arrows denote forward computation (black), local updates (red), and stop-gradient paths (slashed).}

	\label{fig:svp_overview}
    \vspace{-1em}
\end{figure*}

This work targets mathematical analysis, algorithmic development, and experimental evaluations, leading to three principal contributions:
\textbf{Theoretical contribution}: we formally decompose the network ELBO into layer-wise terms under a Markov assumption and prove that the arithmetic mean of these layer-wise ELBOs provides a valid lower bound on the global ELBO, establishing the theoretical basis for local training.
\textbf{Algorithmic contribution}: we proposed SLL and demonstrate its potential as a scalable and efficient alternative to BP. By integrating stochastic random projections, SLL replaces the need for a complete backward pass, thereby facilitating structured local learning. 
\textbf{Experimental evaluations}: We demonstrate that SLL scales effectively across architectures and datasets, from MLPs on MNIST to ViTs on ImageNet.  
Our results show that the SLL algorithm surpasses recently proposed local training methods that address the update locking problem of BP. Moreover, SLL approaches or equals the accuracy performance of BP but with a significant reduction in memory ($4$× or more). 

\footnotetext{Code available at: \url{https://github.com/byin-cwi/SSL}}

\section{Background}

% \subsection{Neural Networks and Backpropogation}

In supervised learning tasks, such as classification applications or regression, neural networks are designed to construct mappings between given input data $X$ and the corresponding target label $Y$. Traditional feedforward neural networks have a sequential structure in which each layer processes the output of the previous layer through a parameterized function. Following the classical formulation~\cite{rumelhart1986learning}, such a $L$-layer neural network can be expressed as a chain of its parameterized sub-functions: 
\begin{equation}
    f_{1:L}(x) := f(f(...f(x, \theta_1)..., \theta_{L-1}), \theta_L) \label{eq1:f}
\end{equation}
where $\theta_i\in\Theta$ represents a set of learnable parameters at layer $i$. This hierarchical or Markov structure introduces a sequence of hidden representations $\mathcal{H} = [h_1, h_2, ..., h_L]$ where each representation is defined recursively as $h_i = f(h_{i-1},\theta_i)$. Given the stacked structure of neural networks, each layer builds on the representation of the previous layer. This structure induces a hierarchical representation where higher layers encode increasingly abstract and task-relevant features. 
% \textcolor{red}{In the following, we review both the deterministic and probabilistic perspective on neural network optimization, together forming the foundation of proposed method. }
% \subsection{Backpropogation} 

\textbf{Backpropagation} is the standard approach for network training, aiming to optimize the parameters $\Theta$ of the network given a dataset of input-label pairs $(x,y)$ and a task-relevant loss function $\mathcal{L}(h_L,y)$. During training, input data is propagated through the entire network to generate predictions. The loss function then evaluates the network performance by quantifying the distance between these predictions and labels. Next, BP computes the gradient of the loss with respect to each parameter by recursively applying the chain rule in reverse through the network. The update rule for the parameters at layer $i$, $\theta_i$, are updated iteratively using gradient descent:
\begin{equation}
\theta_i' = \theta_i + \eta \Delta \theta_i; \quad
\Delta \theta_i = \frac{\partial \mathcal{L}}{\partial \theta_i} 
= \frac{\partial \mathcal{L}}{\partial h_i} \cdot \frac{\partial h_i}{\partial \theta_i} 
= \textcolor{blue}{\frac{\partial \mathcal{L}}{\partial h_L} \prod_{j>i} \frac{\partial h_{j+1}}{\partial h_j}} 
\cdot \textcolor{red}{\frac{\partial h_i}{\partial \theta_i}}
\end{equation}
where $\eta$ is the learning rate. 
The first term (blue) captures the global contributions of activation $h_i$ to the global loss. It encodes dependencies across all subsequent layers and ensures that updates are coordinated with the global objective. The second term (red) reflects the local sensitivity of $h_i$ with respect to the corresponding parameters $\theta_i$, and can be calculated independently at each layer.

\section{Methodology}

In this section, we break the global training objective into local layer updates, so each layer learns locally while still contributing to the overall optimization of the network.

\subsection{From global loss to global ELBO}
%{From Forward Propagation to Variational Inference}
In principle,  BP's inefficiencies arise from its treatment of activations as fixed, deterministic values that require explicit gradient computations across all layers. 
Here, we adopt a probabilistic formulation where each hidden activation is modeled as stochastic latent variables, conditioned on its previous layer. This hierarchy views forward computation as an approximate inference over latent variables, similar to the approaches in deep-generative models~\cite{kingma2014auto, sonderby2016ladder}.
% Here, we adopt a probabilistic perspective to view the deep network as hierarchical latent-variable models, in which activations are interpreted as stochastic variables that transform input distributions at different levels of abstraction~\cite{kingma2014auto, sonderby2016ladder}. 
Thus, instead of optimizing deterministic activations, learning becomes an inference problem where the goal is to infer their posterior distributions conditioned on observed inputs and outputs. Formally, this corresponds to estimating the \textit{true posterior} over the hidden representations:
\begin{equation}
p(h_1,\dots,h_L \mid x,y) = \frac{p(y \mid h_L) p(h_L \mid h_{L-1}) \dots p(h_1 \mid x)}{p(y \mid x)} = \prod_{i=1}^{L+1} p(h_i \mid h_{i-1}) / p(y|x) \label{eq1} \tag{Assumption 1}
\end{equation}
where $h_0 := x$ and $h_{L+1} := y$. This joint distribution factorizes into a global \textit{evidence} term and a product of local conditional terms. However, computing the evidence term requires marginalization over all hidden representations:
$
p(h\mid x,y) = \int \dots \int \prod_{i=1}^{N+1} p(h_i \mid h_{i-1}) \, dh_L \dots dh_1
$
which is computationally intractable in high-dimensional deep architecture.

To address this challenge, we apply Variational Inference (VI)~\cite{blei2017variational,ranganath2014black} to approximate the intractable true posterior $p(y \mid x)$ with a variational surrogate distribution $q(h)$ by minimizing the KL divergence between them in latent space:
$$KL(q(h) \| p(h \mid x)) = \mathbb{E}_q [\log q(h)] - \mathbb{E}_q [\log p(h \mid x)].$$ where $\mathbb{E}_q[\cdot]$ denotes expectation under the variational posterior $q(h)$. This leads to maximizing the Evidence Lower Bound (ELBO): \begin{eqbox}
\begin{equation}
\arg\max_{\theta} \mathcal{E}= \mathbb{E}_q [\log p(y \mid h)] - KL(q(h) \| p(h))
\end{equation}
\end{eqbox}
where $p(h)$ is the prior distribution over latent variables. 
At this point, network optimization is reformulated as a structured variational inference problem, fundamentally distinct from standard BP.

\subsection{From global ELBO to Layer-wise ELBO}

\paragraph{Generative and recognition models.}
We view the network as a hierarchical latent variable model with generative transitions $p(h_i\mid h_{i-1})$ for $i=1,\dots,L$ and likelihood $p(y\mid h_L)$. 
To approximate the intractable posterior, we adopt a Markov assumption on the network architecture that mirrors the forward architecture~\cite{vahdat2020nvae}:
\begin{equation}
q(h_1,\dots,h_L \mid x,y) \;=\; \prod_{i=1}^{L} q(h_i \mid h_{i-1}),
\label{eq2} \tag{Assumption 2}
\end{equation}
where each factor may include auxiliary noise (reparameterization) or reduce to a delta, as specified below. 
Here $p(h_i\mid h_{i-1})$ denotes the generative transition (prior) at layer $i$, and $q(h_i\mid h_{i-1})$ is the approximate posterior (inference distribution) over $h_i$ given $h_{i-1}$. Under this factorization, a standard network-level variational objective is
\begin{equation}
\mathcal{E}_{NN}
\;=\;
\mathbb{E}_{q}\!\big[\log p(y \mid h_L)\big]
\;-\;
\sum_{i=1}^{L} \mathrm{KL}\!\big(q(h_i \mid h_{i-1}) \,\|\, p(h_i \mid h_{i-1})\big),
\label{eq:enn} \tag{Assumption 3}
\end{equation}
with expectation over $q(h_1,\dots,h_L \mid x,y)$. 
Each additive item admits a local interpretation, motivating the following layer-wise ELBO-inspired objective:
\begin{eqbox}
\begin{equation}
\mathcal{E}_{i}
\;=\;
\underbrace{\mathbb{E}_{q(h_{i}\mid x,y)}\!\left[\log p(y \mid h_i)\right]}_{\text{Expected log-likelihood}}
\;-\;
\underbrace{\mathrm{KL}\!\big(q(h_i \mid h_{i-1}) \,\|\, p(h_i \mid h_{i-1})\big)}_{\text{Layer-wise divergence}} ,
\end{equation}
\end{eqbox}
where the first term encourages class-discriminative representations at layer $i$, and the second term regularizes by enforcing local consistency with $p(h_i \mid h_{i-1})$. In short, each layer learns to improve the prediction while remaining consistent with its generative prior~\cite{eldan2016power}.

\paragraph{Layer-to-network relation.}
\textbf{Theorem 1.} \textit{Under the above assumptions, the arithmetic mean of the $L$ layer-wise objectives provides a lower bound surrogate that is dominated by the network objective:}
$
\frac{1}{L}\sum_{i=1}^{L} \mathcal{E}_i \;\le\; \mathcal{E}_{NN}.
$
\textit{Proof sketch in Appendix.}
This result ensures that local optimization at each layer contributes meaningfully to the global objective, thereby supporting SLL as a practical alternative to backpropagation.

% \textbf{Implementation note.} In our implementation, we use a \emph{deterministic} recognition map,
% $q(h_i \mid h_{i-1})=\delta\!\big(h_i - f_i(h_{i-1})\big)$ and compute the layer-wise divergence on auxiliary categorical summaries defined in the next subsection.

\subsection{Stochastic Layer-wise Learning (SLL)}

To approximate the layer-wise ELBO in Assumption 3 with a strictly local training rule, we make each layer-wise KL term as a tractable surrogate defined on auxiliary discrete posteriors coming from adjacent layers. For layer $i$, we attach a random lightweight classification head $R_i:\mathbb{R}^{d_i}\!\to\!\mathbb{R}^{K}$ and define two categorical distributions over $K$ codes induced from the activations: a predictive prior $p_i(\cdot\mid h_{i-1}^{\mathrm{sg}})=\mathrm{softmax}(R_{i-1}h_{i-1}^{\mathrm{sg}})$ that depends only on the stop-gradient parent $h_{i-1}^{\mathrm{sg}}$ (i.e. frozen input from the previous layer), and an auxiliary posterior $q_i(\cdot\mid h_i)=\mathrm{softmax}(R_i h_i)$ which depends on the current activations $h_i$. We replace $\mathrm{KL}\big(q(h_i\mid h_{i-1})\|p(h_i\mid h_{i-1})\big)$, the KL term in the ELBO, by the per-sample Bhattacharyya surrogate:
$$
\mathcal{L}^{\mathrm{per}}_{\mathrm{BC}}(i)
= -\frac{1}{B}\sum_{b=1}^{B}\log \mathrm{BC}\big(q_i^{(b)},p_i^{(b)}\big),
\qquad 
\mathrm{BC}(u,v)=\sum_{k=1}^{K}\sqrt{u_k v_k}\in[0,1].
$$
Here $\mathrm{BC}$ denotes the \emph{Bhattacharyya coefficient}, introduced by Bhattacharyya ~\cite{bhattacharyya1943} as a measure of affinity between distributions; it equals the inner product of square-rooted probabilities. It is closely related to the squared Hellinger distance, since $H^2(u,v)=1-\mathrm{BC}(u,v)$~\cite{bhattacharyya1943,vanErvenHarremoes2014}. This construction preserves locality because $p_i$ depends only on the frozen inputs $h_{i-1}^{\mathrm{sg}}$, while also serving as a proxy for the ELBO term. A second-order expansion yields $\mathrm{KL}(q\|p)=4\big(1-\mathrm{BC}(q,p)\big)+o(\|q-p\|^2)$. Moreover, the inequalities $\mathrm{KL}(q\|p)\ge -2\log \mathrm{BC}(q,p)\ge 2\big(1-\mathrm{BC}(q,p)\big)$ provide global monotone control and improved numerical stability, especially when probabilities are small. The resulting layer objective becomes:
\begin{eqbox}
\begin{equation}
\arg\min_{\theta}\, \mathcal{L}_i
=  \mathcal{L}_{\text{pred}} + \mathcal{L}_{\text{BC}}^{per}
=
\underbrace{\mathcal{L}_{\mathrm{pred}}(R_i h_i, y)}_{\text{expected likelihood term}}
+
% \underbrace{\mathcal{L}^{\mathrm{per}}_{\mathrm{BC}}(i)}_{\text{surrogate for }\mathrm{KL}(q(h_i\mid h_{i-1})\|p(h_i\mid h_{i-1}))},
\underbrace{\mathcal{L}^{\mathrm{per}}_{\mathrm{BC}}(i)}_{\text{surrogate for }\mathrm{KL}(q\|p)},
\end{equation}
\end{eqbox}
is ELBO-inspired rather than a strict ELBO lower bound. In general, optimizing $\{\mathcal{L}_i\}_{i=1}^{L}$ gives a structured approximation to the layer-wise ELBOs in Assumption~3 and, together with Theorem~1, links these local updates to the global objective $\mathcal{E}_{NN}$, thereby enabling scalable training that remains faithful to the hierarchical variational formulation. Unlike auxiliary heads, greedy training, or reconstruction-based target propagation, our local objective is relational across depth which enforces \emph{adjacent-layer probabilistic alignment} by minimizing a Bhattacharyya KL-surrogate between induced posteriors with stop–gradient on the parent, thereby regularizing inter-layer information flow while preserving strict locality.

\paragraph{Stochastic Random Projection.}
We compute layer-wise divergences in a compressed subspace using fixed random projections, which preserve minibatch geometry with high probability  by the JL lemma~\cite{johnson1984extensions}. Concretely, activations are mapped as $v_i=\tfrac{1}{\sqrt{d'}}\,R_i h_i$ with $R_i\in\mathbb{R}^{K\times d}$ sampled once at initialization with i.i.d.\ subgaussian entries, where $K\ll d$ and we set $K$ to the number of classes. For any finite set $\mathcal{H}$ of size $n$ (e.g., a minibatch), the JL lemma ensures that if $d'\!\ge\!C\,\varepsilon^{-2}\log(n/\delta)$ then, with probability at least $1-\delta$, pairwise distances and inner products among $\{v_i(u):u\in\mathcal{H}\}$ are preserved up to $O(\varepsilon)$; this justifies computing our alignment divergence on the auxiliary posteriors in the projected space. The projections act as lightweight heads that enable strictly local updates without backpropagating across layers. To improve generalization, we inject structured noise into the projection during training:
\[
v_i \;=\; \tfrac{1}{\sqrt{d'}}\,(M_i \odot R_i)\,h_i,\qquad M_i\sim\mathrm{Bernoulli}(p)^{d'\times d},
\]
which acts as multiplicative dropout on the projection weights. This introduces Monte Carlo variability without learning the projection, and is consistent with the Bayesian view of dropout as approximate variational inference while our overall objective remains ELBO-inspired~\cite{gal2016dropout}. In our implementation it functions as a stochastic regularizer that stabilizes the induced posteriors and improves robustness. The result is a geometry-preserving, parameter-efficient mechanism that stabilizes alignment, mitigates over-compression, and scales local training.

\textbf{Implementation note.} We use a deterministic approximate posterior $q(h_i\mid h_{i-1})=\delta\!\big(h_i-f_i(h_{i-1})\big)$ and therefore compute the layer-wise divergence on the auxiliary categorical summaries $(q_i,p_i)$ rather than the continuous conditionals, preserving locality via stop–gradient on the prior side.
During training, each layer is updated locally as the child-side distribution $q_i$, while its frozen output simultaneously serves as the parent-side target $p_{i+1}$ for the next layer, yielding a chain of coordinated adjacent-layer updates without cross-layer backpropagation.

\section{Related Work}

The intersection of probabilistic inference and biologically plausible optimization has inspired a range of methods that seek to improve the scalability, interpretability, and local adaptability of deep learning. We organize related work into three areas: variational inference, local learning, and forward-only training.
\textbf{Variational Inference and Probabilistic Deep Learning.} 
Variational inference (VI) enables tractable approximate Bayesian learning via ELBO maximization~\cite{blei2017variational,jordan1999introduction}, foundational to deep generative models like VAEs~\cite{kingma2014auto, sohn2015learning, higgins2017beta}, and their structured extensions~\cite{sonderby2016ladder, vahdat2020nvae}. SLL approximates VI for feedforward networks, combining local latent approximations with task-driven learning, and can be seen as a layer-wise variational EM scheme.

\textbf{Gradient-Based Local Learning}
Local learning reduces backpropagation overhead by optimizing layers independently, from greedy layer-wise training~\cite{bengio2006greedy} to local heads~\cite{belilovsky2019greedy, nokland2019training} and synchronization strategies~\cite{ernoult2022towards}. Recent blockwise and parallel approaches~\cite{yang2024towards, apolinario2024lls} aim to scale under memory constraints, but often suffer from global feature inconsistency~\cite{yang2024towards}. SLL alleviates this via variational alignment. More biologically inspired alternatives include FA~\cite{lillicrap2016random}, DFA~\cite{nokland2016direct}, DPK~\cite{webster2021learning}, and TP~\cite{lee2015difference}, which replace gradients with alternative feedback signals. More recent Hebbian variants~\cite{journe2023hebbian, halvagal2023combination} show promise for scalable bio-plausible learning, though accuracy and depth remain challenges.

\textbf{Forward-Only Credit Assignment} 
Forward-only methods eliminate backprop by using dual forward passes, e.g., Forward-Forward (FF)\cite{hinton2022forward}, Signal Propagation\cite{kohan2023signal}, and PEPITA~\cite{dellaferrera2022error}. Other FF variants~\cite{wu2024distance, dooms2023trifecta, lee2023symba} reframe credit assignment via $L_2$ distances. Despite biological inspiration, these methods often face inter-layer misalignment~\cite{lorberbom2024layer}, limiting hierarchical feature learning.

\begin{figure*}[!t]
	\centering
    \includegraphics[width=.95\textwidth]{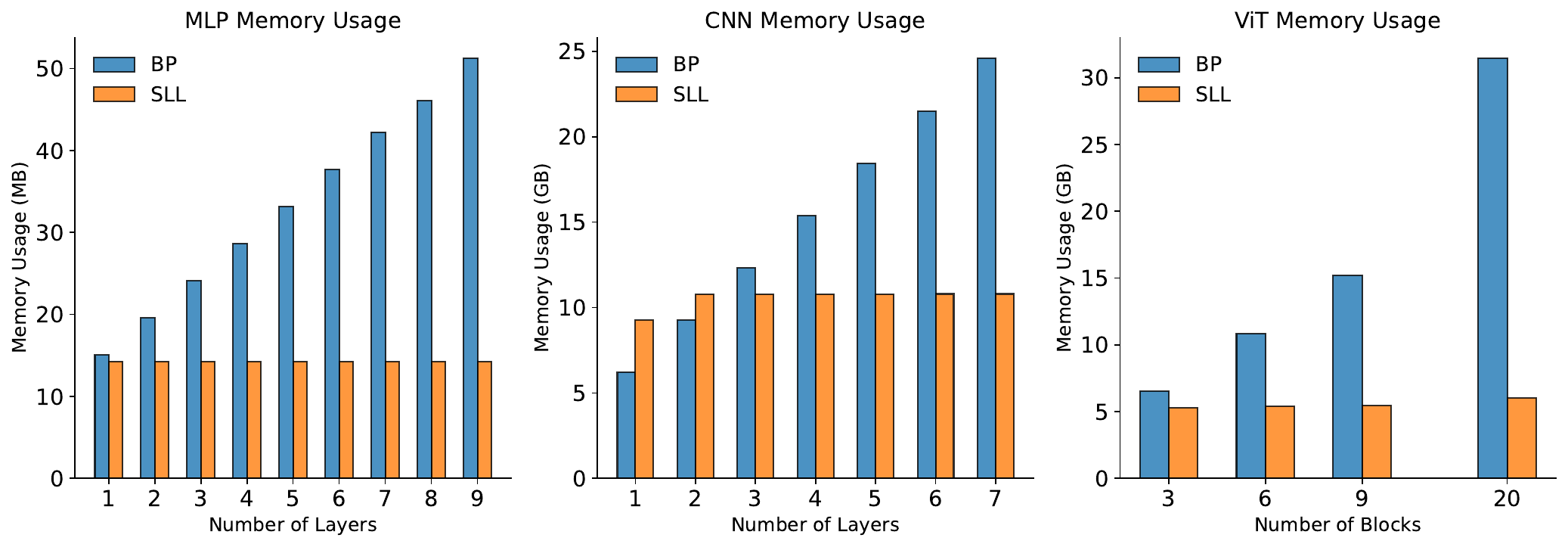}
	\caption{Peak training memory on (a) MLPs (1024 neurons/layer) as a function of depth. BP memory scales linearly, while SLL remains constant; (b) CNNs on the Imagenette  without pooling layers. Each convolution layer uses a kernel size of 3 and 64 output channels; (c) ViTs on Imagenette. For fair comparison, we are using SGD as the optimizer in training.}
    \vspace{-10pt}
	\label{fig:mems-mlp-cnn}
\end{figure*}

\section{Experiments}

We evaluate the effectiveness, interpretability, and scalability of SLL across a range of standard benchmarks. Our experiments include multiple architectures, including MLPs, CNNs, and Vision Transformers (ViTs), and datasets of increasing complexity, from MNIST~\cite{lecun1998gradient} and CIFAR-10/100~\cite{krizhevsky2009learning} to ImageNette and ImageNet-1K~\cite{deng2009imagenet}. To assess SLL’s capacity for local learning, we compare it against established local training baselines across multiple network scales. We further extend SLL to block-wise training (SLL+) for ViTs, demonstrating its compatibility with modern large-scale architectures without relying on full backpropagation.

\begin{table}[!h]

    \centering

    \resizebox{!}{1.76cm}{  % width = auto (!), height = 5cm
    \begin{tabular}{lllccc}
        \hline
        \textbf{Method} & \textbf{Memory} & \textbf{FLOPS}  & \textbf{MNIST} & \textbf{CIFAR10} & \textbf{CIFAR100} \\ \hline
        \midrule
        BP &$\mathcal{O}(NL)$&$\mathcal{O}(N^2L)$ &99.25 $\pm$ 0.09 &60.95 $\pm$ 0.33 &32.92 $\pm$ 0.23 \\
        \hline
        % BP &$\mathcal{O}(NL)$&$\mathcal{O}(N^2L)$ & CNN&99.62 $\pm$ 0.02 &87.57 $\pm$ 0.13 &62.25 $\pm$ 0.29 \\ \hline
        TP  \cite{lee2015difference}&$\mathcal{O}(NL)$&$\mathcal{O}(N^2L)$ & 97.96 $\pm$ 0.08&49.64 $\pm$ 0.26  & -\\
        FA\cite{lillicrap2016random}&$\mathcal{O}(NL)$&$\mathcal{O}(NLC)$ & 98.36 $\pm$ 0.03&53.10 $\pm$ 0.30  & 25.70 $\pm$ 0.20\\
        % FA&$\mathcal{O}(NL)$&$\mathcal{O}(NLC)$ & MLP& 97.90 $\pm$ 0.17&71.53 $\pm$ 0.4  & 44.93 $\pm$ 0.5\\
        DFA\cite{nokland2016direct} &$\mathcal{O}(NL)$&$\mathcal{O}(NLC)$ &98.26 $\pm$ 0.08 &57.10 $\pm$ 0.20 & 26.90 $\pm$ 0.10\\ 
     PEPITA\cite{dellaferrera2022error} &$\mathcal{O}(NL)$&$\mathcal{O}(N^2L)$ &98.01 $\pm$ 0.09 &52.57 $\pm$ 0.36 & 24.91 $\pm$ 0.22\\ 
        SP\cite{kohan2023signal} &$\mathcal{O}(N)$&$\mathcal{O}(N^2L)$ &98.29 $\pm$ 0.03 &57.38 $\pm$ 0.16 & 29.70 $\pm$ 0.19\\ \hline
        SLL&$\mathcal{O}(N)$&$\mathcal{O}(NLC)$ &\textbf{99.32 $\pm$ 0.05} &\textbf{61.43 $\pm$ 0.31} & \textbf{32.95  $\pm$ 0.26} \\
        % SLL&$\mathcal{O}(N)$&$\mathcal{O}(NLC)$ & CNN&\textbf{98.79$\pm$0.05} &\textbf{61.43$\pm$0.31} & \textbf{32.95$\pm$ 0.26} \\
        \hline
        \hline
    \end{tabular}
    }

    \caption{Performance and computational complexity of SLL vs prior local-learning methods for MLPs on MNIST, CIFAR-10, and CIFAR-100 under the same experimental setup. BP and baseline results are taken from \citep{kohan2023signal}. Memory and FLOPs are reported as asymptotic scaling in $N$ (neurons per layer), $L$ (layers), and $C$ (classes). Metrics are mean $\pm$ std over three runs. “–” denotes values not reported.}

    \label{tab:tab1}
\end{table}

\subsection{Experiments on MLPs}

We begin by evaluating SLL on fully connected networks trained on benchmarks: MNIST and CIFAR-10/100. These datasets serve as controlled settings to study local learning dynamics in low-dimensional and moderately complex inputs.

\begin{figure*}[h]
	\centering
    \includegraphics[width=.95\textwidth]{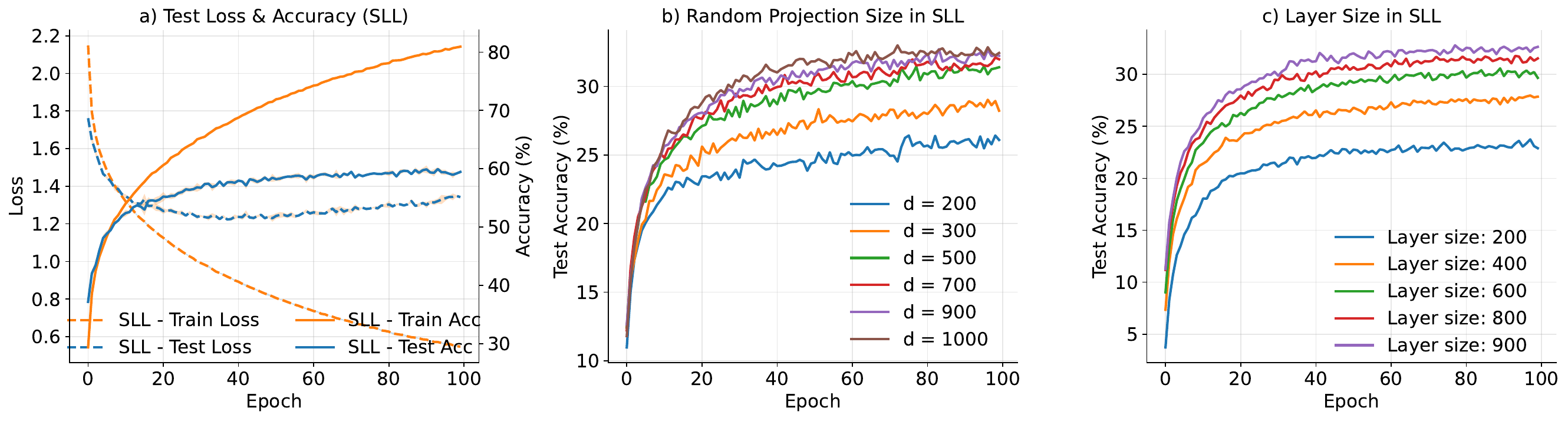}
	\caption{(a) Training curves of a 3-layer MLP on CIFAR-10 via SLL. Ablation study: (b) random projection dimension in a 3×1000 MLP trained on CIFAR-100; activations are downsampled to $d$-dim via adaptive pooling before projection. (c) network width in SLL on CIFAR-100, showing that wider layers significantly enhance performance and stability.}
    \vspace{-7pt}
	\label{fig:abl_sum}
\end{figure*}

\textbf{Accuracy and Efficiency.}
To establish a comprehensive comparison, we evaluate SLL alongside a range of biologically motivated and local learning algorithms that do not rely fully or avoid BP. All models are trained with identical architectures and training schedules to ensure a fair comparison. As shown in Table~\ref{tab:tab1} and Figure~\ref{fig:abl_sum}(a), SLL consistently outperforms all local learning baselines, despite operating under reduced memory and computational budgets. In particular, under this identical setting in ~\cite{kohan2023signal}, SLL even surpasses BP on these datasets while requiring fewer operations and avoiding global gradient synchronization. Moreover, Figure~\ref{fig:mems-mlp-cnn}(a) confirms SLL’s memory efficiency during training. The training memory usage of SLL remains effectively constant as the depth of the network increases, in contrast to its theoretical complexity reported in Table~\ref{tab:tab1}. 

\begin{figure}[!htb]
	\centering
    \includegraphics[width=.75\textwidth]{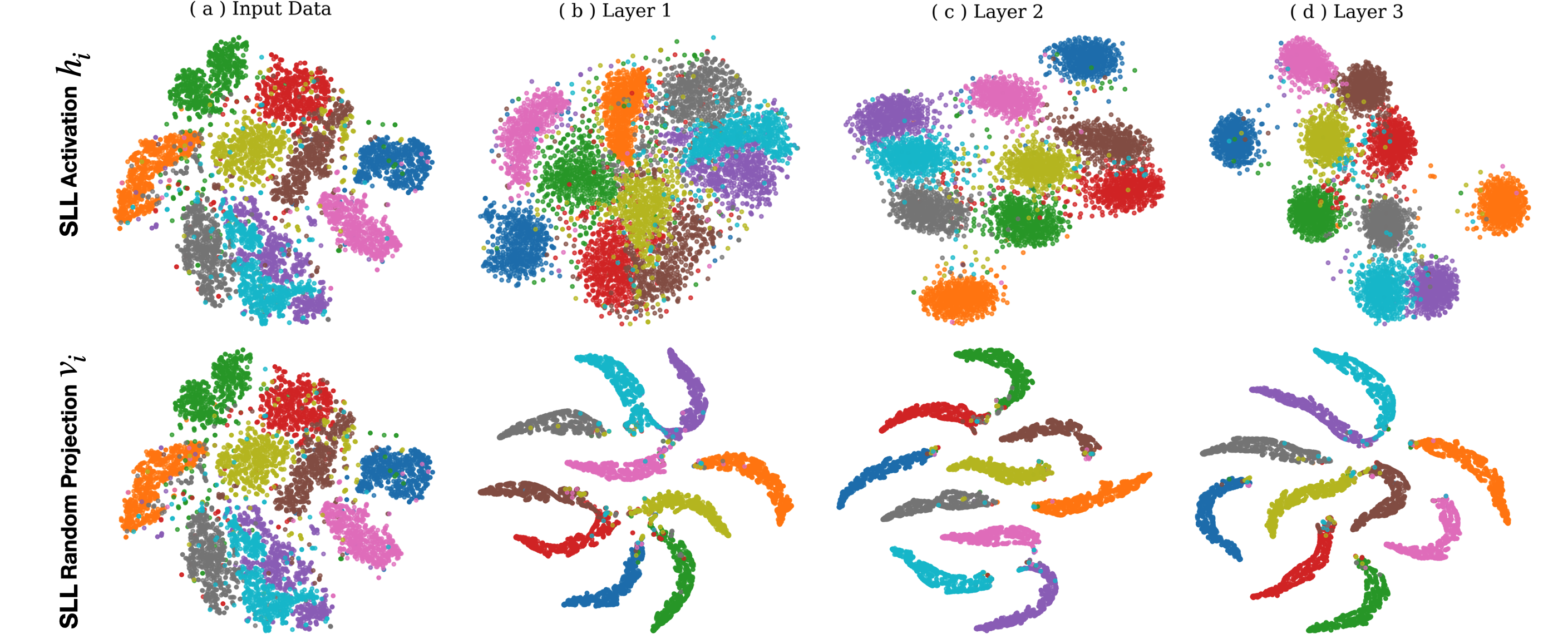}
	\caption{t-SNE visualization of activations and random projections on MNIST, colored by class.}
	\label{fig:tsne-mlp-c10}
\end{figure}

\textbf{Representation Visualization.}
We analyze the internal representations of the network trained by SLL in Figure~\ref{fig:tsne-mlp-c10}. In general, input features are initially entangled, deeper layers show improved class separation. It is obvious that $v_i$ forms sharper, more distinct clusters than $h_i$, indicating that random projections not only preserve but often enhance class-discriminative structure.

\textbf{Ablation study.} We further investigate the effect of projection dimension and network width on SLL performance (Figure~\ref{fig:abl_sum}b,c).
Increasing the projection dimension $d$ improves test accuracy, with diminishing returns beyond $d = 700$, suggesting a trade-off between representational precision and efficiency. Likewise, wider networks result in faster convergence and higher accuracy on CIFAR-100, with improvements saturating above 800 neurons. These trends are consistent with our theoretical insights in JL Lemma~\cite{johnson1984extensions}, which indicate that high-dimensional layers reduce alignment loss and preserve inter-layer information. Together, these findings highlight the role of capacity and compression in enabling stable local learning with SLL.

\subsection{Scaling SLL to CNNs}

We next explore how SLL can scale effectively to convolutional architectures despite discarding explicit spatial structure when utilizing fully connected random projection. To this end, we evaluate SLL on a VGG-11 architecture and compare it against representative local learning methods, forward-only training algorithms, and conventional global BP. 

\textbf{Accuracy.} Table~\ref{tab:cnn_omparison} reports the test accuracies in F-MNIST, CIFAR-10/100 and Tiny-Imagenet. SLL performs competitively with BP, achieving within 1--2\% of BP on all datasets like F-MNIST, CIFAR-10/100 and TinyImageNet200, even slightly surpassing it on F-MNIST. In particular, SLL outperforms all local and forward-only baselines on all given tasks, including DFA~\cite{nokland2016direct}, DKP~\cite{webster2021learning}, SoftHebb~\cite{journe2023hebbian}, and TFF~\cite{dooms2023trifecta}.

\begin{table}[!t]
    \centering
    % \begin{tabular}{lccccc}
    \begin{tabular*}{\linewidth}{@{\extracolsep{\fill}} l c c c c c @{}}
        \hline
        Model & F-MNIST & CIFAR10 & CIFAR100 & Imagenette & Tiny-Imagenet$_{64}$ \\
        \hline
        
        BP-CNN & 93.52(0.22) & 91.58(0.53) & 68.7(0.38) & 90.5(0.45) & 48.15(0.82)\\
        \hline    
        \multicolumn{5}{l}{\textbf{Local Learning}} \\
        FA  \cite{nokland2016direct} & 91.12(0.39) & 60.45(1.13) & 19.49(0.97) & -- & --\\
        DFA \cite{nokland2016direct} & 91.54(0.14) & 62.70(0.36) & 48.03(0.61) & -- & 32.12(0.66)\\
        DKP \cite{webster2021learning} & 91.66(0.27) & 64.69(0.72) & 52.62(0.48) & -- & 35.37(1.92)\\
        Softhebb \cite{journe2023hebbian} & -- & 80.3 & 56 & 81.0 & --\\
        SGR \cite{yang2024towards} & -- & 72.40(0.75) & 49.41(0.44) & -- & --\\
        LLS \cite{apolinario2024lls} & 90.54(0.23) & 88.64(0.12) & 58.84(0.33) & -- & 35.99(0.38)\\
        \hline
        \multicolumn{5}{l}{\textbf{Forward-Only}} \\
        FF-CNN \cite{hinton2022forward} & -- & 59 & -- & -- & --\\
        TFF \cite{dooms2023trifecta} & 91.44(0.49) & 83.51(0.78) & 35.26(0.23) & -- & --\\
        PEPITA \cite{dellaferrera2022error} & -- & 56.33(1.35) & 27.56(0.60) & -- & -- \\
        LC-FF \cite{lorberbom2024layer} & 88.4 & 48.4 & -- & -- & --\\
        DF-R \cite{wu2024distance} & 92.5 & 84.75 & 48.16 & 81.2 & --\\
        \hline
        \textbf{SLL-CNN} & \textbf{93.67(0.17)} & \textbf{91.36(0.32)} & \textbf{67.57(0.18)} & \textbf{88.09(0.73)} & \textbf{49.42(0.65)}\\
        \bottomrule
    \end{tabular*}
    % \caption{Comparison of test accuracies across various datasets, values are mean(std) over 3 runs.}
    \caption{CNN test accuracies comparing SLL with prior local-learning and forward-only methods. Values are reported as mean(std) over three runs; “–” indicates not reported.}

    \vspace{-10pt}
    \label{tab:cnn_omparison}
\end{table} 

\begin{figure}[!t]
	\centering
	\includegraphics[width=.95\textwidth]{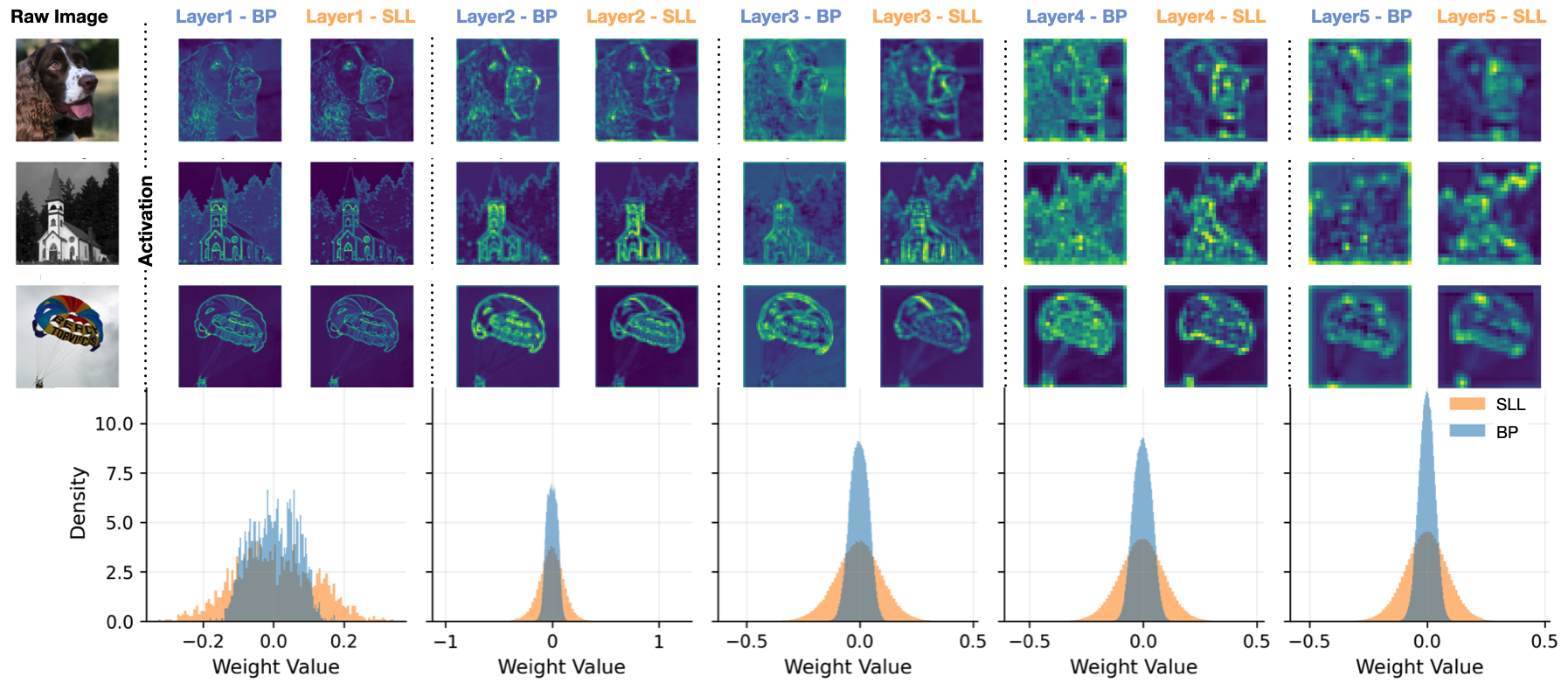}
    \caption{Activation and weight distributions from VGG-11 trained with BP and SLL on Imagenette.}
    \vspace{-15pt}
	\label{fig:c10_vis}
\end{figure}
\textbf{Training memory efficiency.}
Figure~\ref{fig:mems-mlp-cnn}(b) illustrates the training memory usage of SLL and BP on CNNs. While SLL exhibits a clear memory advantage in MLPs, its benefit is more moderate in CNNs. This is because convolution operations are inherently sparse and memory-efficient, while the dense random projections used in SLL introduce additional overhead. However, SLL still maintains a significant advantage in deeper architectures. 

\textbf{Feature visualization.} Figure~\ref{fig:c10_vis} indicates that SLL effectively learns high-quality spatial and discriminative representations, despite discarding explicit spatial priors. Compared with BP, the broader weight distributions from SLL suggest robust and distributed encoding. 

\vspace{23pt}

\subsection{Scaling to Vision Transformer}

\begin{wraptable}{r}{0.59\textwidth}
    \vspace{-16pt}
    \centering
    \begin{tabular}{llcc}
        \toprule
        Task                 & Method   & Test Acc & Memory(GB) \\
        \midrule
        \multirow{2}{*}{CIFAR-10} & BP       & 93.62 &  3.05\\
                             & SLL$^{7+}$      & 92.17 &  \textbf{1.18($\downarrow$ 64.1\%)}\\
        \midrule
        \multirow{2}{*}{CIFAR-100} & BP      & 75.24 &  3.05\\
                              & SLL$^{7+}$     & 74.27 &  \textbf{1.18($\downarrow$ 64.1\%)}\\
        \midrule
        \multirow{2}{*}{Imagenette}& BP & 92.82 &  22.12\\
                                  & SLL$^{7+}$ & 92.25 &  \textbf{5.43($\downarrow$ 75.45\%)}\\
        \midrule
        \multirow{4}{*}{Imagenet} & BP\cite{yuan2021tokens}  & 79.4  & 20.70 \\
                                  & SGR\cite{yang2024towards}$^{3+}$  & 78.65 & 11.73($\downarrow$ 43.33\%) \\
                                  & SLL$^{3+}$ & 72.43 &  \textbf{6.54($\downarrow$ 68.41\%)}\\
                                  & SLL$^{12+}$ & 59.62 &  \textbf{4.30($\downarrow$ 79.22\%)}\\
        \bottomrule
    \end{tabular}
        % \vspace{-9pt}   
    % \caption{Experimental results. 'Memory' refers to GPU training memory usage with batch sizes of 128 and 256 on ImageNet. SGR is \cite{yang2024towards} and BP base line is from \cite{yuan2021tokens}}
    % \caption{Experimental results on ImageNet. \emph{Memory} reports peak GPU training memory for batch sizes $128$ and $256$. SGR results are from \citep{yang2024towards}; the BP baseline follows \citep{yuan2021tokens}.}
    \caption{ViTs results. “Memory” denotes peak GPU training memory at batch sizes 128 and 256.}
    % \textcolor{red}{update}}
    % For SLL+, peak memory usage is reported.\textcolor{red}{update}}
    % \vspace{-10pt}
    \label{tab:vit}

% \end{table}
\end{wraptable}
Moreover, we use Vision Transformers 
(ViTs)~\cite{dosovitskiy2020image} as a scalability benchmark for SLL, since their dense, MLP-like blocks and large activations footprints heavily impact compute and memory, making them ideal for testing efficiency and convergence.

To scale to ViTs, we propose SLL$^{i+}$, a blockwise variant of SLL tailored for large residual architectures. We partition the ViT architecture into $i$-units, each comprising one or more attention blocks; training is hybrid where standard backpropagation is used within each unit, while between units we optimize the local objectives independently, eliminating global backpropagation across the entire model. It effectively turns SLL into a local block-wise training scheme for deep networks, in this case ViTs.
This design aligns with the residual structure of ViT while preserving the localized memory and learning advantage of SLL. 

SLL$^{i+}$ leverages the class token or mean over all tokens as a stable and semantically meaningful signal for local supervision. This allows efficient classification without requiring end-to-end backpropagation. As shown in table~\ref{tab:vit}, SLL$^{i+}$ achieves large memory savings in Vision Transformers while preserving accuracy, with memory use staying nearly constant as block depth increases (Figure~\ref{fig:mems-mlp-cnn}(c)).
%demonstrates accuracy and peak memory usage of SLL$^{i+}$. SLL$^{i+}$ maintains substantial memory savings in ViTs, while preserving accuracy, with memory usage remaining nearly constant with increasing block depth (Figure~\ref{fig:mems-mlp-cnn}(c)). 
This trend is similar to the MLP findings and demonstrates SLL scalability across architectures. Compared to BP, SLL$^{i+}$ reduces training memory by 64\%–80\% without sacrificing stability or model capacity. %making it especially suitable for deep or resource-limited settings.

\section{Discussion and Conclusions}
The above results highlight open opportunities for improving SLL. First, the Markov assumption between layers, while simplifying inference, may limit expressivity in architectures with long-range dependencies such as residual connections. Second, the absence of second-order gradient information may reduce SLL's effectiveness in navigating ill-conditioned loss surfaces. Third, SLL's reliance on local supervision may limit convergence in large-scale classification tasks where informative gradients may only emerge in later layers. Finally, aggressive dimension reduction via random projection may lead to information loss in narrow architectures. Addressing these challenges through more expressive dependency modeling, adaptive projection schemes, architecture-aware supervision, and specialized training approaches for sequential models could extend the applicability of SLL to broader research.

It is worth mentioning that the SLL also draws conceptual parallels with Equilibrium Propagation (EP)~\cite{scellier2017equilibrium} and energy-based models. Both frameworks enable local updates that align with global objectives, but they operate through distinct mechanisms: stochastic layer-wise updates for SLL and dynamical relaxation for EP. Bridging these perspectives under a unified probabilistic or dynamical systems framework is an interesting direction for future research. 

In conclusion, we introduce SLL, a scalable and memory-efficient alternative to BP that reformulates training as an ELBO inspired, stochastic layer-wise learning.  By combining stochastic random projection with a Bhattacharyya surrogate for the layer-wise KL, SLL enables parallel, local updates while preserving global coherence without global BP and without additional trainable parameters. Compared to BP, SLL achieves competitive accuracy with significant memory efficiency, up to 4× in our settings, and consistently outperforms prior local learning methods. It generalizes effectively across MLPs, CNNs, and ViTs, scaling from small to moderately large vision tasks. Beyond training efficiency, SLL also provides a structured probabilistic view of deep representations, offering a foundation for interpretable learning dynamics and architecture design grounded in information flow.

\vspace{\baselineskip}        % ≈ one empty line
\vspace{\baselineskip}        % ≈ one empty line
\vspace{\baselineskip}        % ≈ one empty line
\vspace{\baselineskip}        % ≈ one empty line
% \vspace{\baselineskip}        % ≈ one empty line
% \bibliography{iclr2026_conference}
% \bibliographystyle{iclr2026_conference}

\bibliographystyle{unsrt}  
\bibliography{references}  

\begin{thebibliography}{10}

\bibitem{rumelhart1986learning}
David~E Rumelhart, Geoffrey~E Hinton, and Ronald~J Williams.
\newblock Learning representations by back-propagating errors.
\newblock {\em nature}, 323(6088):533--536, 1986.

\bibitem{lecun2015deep}
Yann LeCun, Yoshua Bengio, and Geoffrey Hinton.
\newblock Deep learning.
\newblock {\em nature}, 521(7553):436--444, 2015.

\bibitem{lillicrap2020backpropagation}
Timothy~P Lillicrap, Adam Santoro, Luke Marris, Colin~J Akerman, and Geoffrey Hinton.
\newblock Backpropagation and the brain.
\newblock {\em Nature Reviews Neuroscience}, 21(6):335--346, 2020.

\bibitem{jaderberg2017decoupled}
Max Jaderberg, Wojciech~Marian Czarnecki, Simon Osindero, Oriol Vinyals, Alex Graves, David Silver, and Koray Kavukcuoglu.
\newblock Decoupled neural interfaces using synthetic gradients.
\newblock In {\em International conference on machine learning}, pages 1627--1635. PMLR, 2017.

\bibitem{griewank2008evaluating}
Andreas Griewank and Andrea Walther.
\newblock {\em Evaluating derivatives: principles and techniques of algorithmic differentiation}.
\newblock SIAM, 2008.

\bibitem{luo2024efficient}
Xiangzhong Luo, Di~Liu, Hao Kong, Shuo Huai, Hui Chen, Guochu Xiong, and Weichen Liu.
\newblock Efficient deep learning infrastructures for embedded computing systems: A comprehensive survey and future envision.
\newblock {\em ACM Transactions on Embedded Computing Systems}, 24(1):1--100, 2024.

\bibitem{belilovsky2019greedy}
Eugene Belilovsky, Michael Eickenberg, and Edouard Oyallon.
\newblock Greedy layerwise learning can scale to imagenet.
\newblock In {\em International conference on machine learning}, pages 583--593. PMLR, 2019.

\bibitem{bengio2006greedy}
Yoshua Bengio, Pascal Lamblin, Dan Popovici, and Hugo Larochelle.
\newblock Greedy layer-wise training of deep networks.
\newblock {\em Advances in neural information processing systems}, 19, 2006.

\bibitem{scellier2017equilibrium}
Benjamin Scellier and Yoshua Bengio.
\newblock Equilibrium propagation: Bridging the gap between energy-based models and backpropagation.
\newblock {\em Frontiers in computational neuroscience}, 11:24, 2017.

\bibitem{guerguiev2017towards}
Jordan Guerguiev, Timothy~P Lillicrap, and Blake~A Richards.
\newblock Towards deep learning with segregated dendrites.
\newblock {\em elife}, 6:e22901, 2017.

\bibitem{whittington2019theories}
James~CR Whittington and Rafal Bogacz.
\newblock Theories of error back-propagation in the brain.
\newblock {\em Trends in cognitive sciences}, 23(3):235--250, 2019.

\bibitem{sacramento2018dendritic}
Jo{\~a}o Sacramento, Rui Ponte~Costa, Yoshua Bengio, and Walter Senn.
\newblock Dendritic cortical microcircuits approximate the backpropagation algorithm.
\newblock {\em Advances in neural information processing systems}, 31, 2018.

\bibitem{yang2024towards}
Yibo Yang, Xiaojie Li, Motasem Alfarra, Hasan Hammoud, Adel Bibi, Philip Torr, and Bernard Ghanem.
\newblock Towards interpretable deep local learning with successive gradient reconciliation.
\newblock In {\em International Conference on Machine Learning}, pages 56196--56215, 2024.

\bibitem{he2023law}
Hangfeng He and Weijie~J Su.
\newblock A law of data separation in deep learning.
\newblock {\em Proceedings of the National Academy of Sciences}, 120(36):e2221704120, 2023.

\bibitem{razdaibiedina2022representation}
Anastasia Razdaibiedina, Ashish Khetan, Zohar Karnin, Daniel Khashabi, and Vivek Madan.
\newblock Representation projection invariance mitigates representation collapse.
\newblock In {\em Findings of the Association for Computational Linguistics: EMNLP 2023}, pages 14638--14664, 2023.

\bibitem{telgarsky2016benefits}
Matus Telgarsky.
\newblock Benefits of depth in neural networks.
\newblock In {\em Conference on learning theory}, pages 1517--1539. PMLR, 2016.

\bibitem{shwartz2017opening}
Ravid Shwartz-Ziv and Naftali Tishby.
\newblock Opening the black box of deep neural networks via information.
\newblock {\em arXiv preprint arXiv:1703.00810}, 2017.

\bibitem{bhattacharyya1943}
Anil~Kumar Bhattacharyya.
\newblock On a measure of divergence between two statistical populations defined by their probability distributions.
\newblock {\em Bulletin of the Calcutta Mathematical Society}, 35:99--109, 1943.

\bibitem{johnson1984extensions}
William~B Johnson, Joram Lindenstrauss, et~al.
\newblock Extensions of lipschitz mappings into a hilbert space.
\newblock {\em Contemporary mathematics}, 26(189-206):1, 1984.

\bibitem{gal2016dropout}
Yarin Gal and Zoubin Ghahramani.
\newblock Dropout as a bayesian approximation: Representing model uncertainty in deep learning.
\newblock In {\em international conference on machine learning}, pages 1050--1059. PMLR, 2016.

\bibitem{kingma2014auto}
Diederik~P. Kingma and Max Welling.
\newblock Auto-encoding variational bayes.
\newblock In {\em International Conference on Learning Representations (ICLR)}, 2014.

\bibitem{sonderby2016ladder}
Casper~Kaae S{\o}nderby, Tapani Raiko, Lars Maal{\o}e, S{\o}ren~Kaae S{\o}nderby, and Ole Winther.
\newblock Ladder variational autoencoders.
\newblock {\em Advances in neural information processing systems}, 29, 2016.

\bibitem{blei2017variational}
David~M Blei, Alp Kucukelbir, and Jon~D McAuliffe.
\newblock Variational inference: A review for statisticians.
\newblock {\em Journal of the American statistical Association}, 112(518):859--877, 2017.

\bibitem{ranganath2014black}
Rajesh Ranganath, Sean Gerrish, and David Blei.
\newblock Black box variational inference.
\newblock In {\em Artificial intelligence and statistics}, pages 814--822. PMLR, 2014.

\bibitem{vahdat2020nvae}
Arash Vahdat and Jan Kautz.
\newblock Nvae: A deep hierarchical variational autoencoder.
\newblock {\em Advances in neural information processing systems}, 33:19667--19679, 2020.

\bibitem{eldan2016power}
Ronen Eldan and Ohad Shamir.
\newblock The power of depth for feedforward neural networks.
\newblock In {\em Conference on learning theory}, pages 907--940. PMLR, 2016.

\bibitem{vanErvenHarremoes2014}
Tim van Erven and Peter Harremo{\"e}s.
\newblock R{\'e}nyi divergence and kullback--leibler divergence.
\newblock {\em IEEE Transactions on Information Theory}, 60(7):3797--3820, 2014.

\bibitem{jordan1999introduction}
Michael~I Jordan, Zoubin Ghahramani, Tommi~S Jaakkola, and Lawrence~K Saul.
\newblock An introduction to variational methods for graphical models.
\newblock {\em Machine learning}, 37:183--233, 1999.

\bibitem{sohn2015learning}
Kihyuk Sohn, Honglak Lee, and Xinchen Yan.
\newblock Learning structured output representation using deep conditional generative models.
\newblock {\em Advances in neural information processing systems}, 28, 2015.

\bibitem{higgins2017beta}
Irina Higgins, Loic Matthey, Arka Pal, Christopher Burgess, Xavier Glorot, Matthew Botvinick, Shakir Mohamed, and Alexander Lerchner.
\newblock beta-vae: Learning basic visual concepts with a constrained variational framework.
\newblock In {\em International conference on learning representations}, 2017.

\bibitem{nokland2019training}
Arild N{\o}kland and Lars~Hiller Eidnes.
\newblock Training neural networks with local error signals.
\newblock In {\em International conference on machine learning}, pages 4839--4850. PMLR, 2019.

\bibitem{ernoult2022towards}
Maxence~M Ernoult, Fabrice Normandin, Abhinav Moudgil, Sean Spinney, Eugene Belilovsky, Irina Rish, Blake Richards, and Yoshua Bengio.
\newblock Towards scaling difference target propagation by learning backprop targets.
\newblock In {\em International Conference on Machine Learning}, pages 5968--5987. PMLR, 2022.

\bibitem{apolinario2024lls}
Marco Paul~E Apolinario, Arani Roy, and Kaushik Roy.
\newblock Lls: local learning rule for deep neural networks inspired by neural activity synchronization.
\newblock {\em arXiv preprint arXiv:2405.15868}, 2024.

\bibitem{lillicrap2016random}
Timothy~P Lillicrap, Daniel Cownden, Douglas~B Tweed, and Colin~J Akerman.
\newblock Random synaptic feedback weights support error backpropagation for deep learning.
\newblock {\em Nature communications}, 7(1):13276, 2016.

\bibitem{nokland2016direct}
Arild N{\o}kland.
\newblock Direct feedback alignment provides learning in deep neural networks.
\newblock {\em Advances in neural information processing systems}, 29, 2016.

\bibitem{webster2021learning}
Matthew~Bailey Webster, Jonghyun Choi, and Changwook Ahn.
\newblock Learning the connections in direct feedback alignment.
\newblock openreview, 2021.

\bibitem{lee2015difference}
Dong-Hyun Lee, Saizheng Zhang, Asja Fischer, and Yoshua Bengio.
\newblock Difference target propagation.
\newblock In {\em Proceedings of the 2015th European Conference on Machine Learning and Knowledge Discovery in Databases-Volume Part I}, pages 498--515, 2015.

\bibitem{journe2023hebbian}
Adrien et~al. Journé.
\newblock Hebbian deep learning without feedback.
\newblock In {\em International conference on learning representations}, 2023.

\bibitem{halvagal2023combination}
Manu~Srinath Halvagal and Friedemann Zenke.
\newblock The combination of hebbian and predictive plasticity learns invariant object representations in deep sensory networks.
\newblock {\em Nature Neuroscience}, 26(11):1906--1915, 2023.

\bibitem{hinton2022forward}
Geoffrey Hinton.
\newblock The forward-forward algorithm: Some preliminary investigations.
\newblock {\em arXiv preprint arXiv:2212.13345}, 2022.

\bibitem{kohan2023signal}
Adam Kohan, Edward~A Rietman, and Hava~T Siegelmann.
\newblock Signal propagation: The framework for learning and inference in a forward pass.
\newblock {\em IEEE Transactions on Neural Networks and Learning Systems}, 2023.

\bibitem{dellaferrera2022error}
Giorgia Dellaferrera and Gabriel Kreiman.
\newblock Error-driven input modulation: solving the credit assignment problem without a backward pass.
\newblock In {\em International Conference on Machine Learning}, pages 4937--4955. PMLR, 2022.

\bibitem{wu2024distance}
Yujie Wu, Siyuan Xu, Jibin Wu, Lei Deng, Mingkun Xu, Qinghao Wen, and Guoqi Li.
\newblock Distance-forward learning: Enhancing the forward-forward algorithm towards high-performance on-chip learning.
\newblock {\em arXiv preprint arXiv:2408.14925}, 2024.

\bibitem{dooms2023trifecta}
Thomas Dooms, Ing~Jyh Tsang, and Jose Oramas.
\newblock The trifecta: Three simple techniques for training deeper forward-forward networks.
\newblock {\em arXiv preprint arXiv:2311.18130}, 2023.

\bibitem{lee2023symba}
Heung-Chang Lee and Jeonggeun Song.
\newblock Symba: Symmetric backpropagation-free contrastive learning with forward-forward algorithm for optimizing convergence.
\newblock {\em arXiv:2303.08418}, 2023.

\bibitem{lorberbom2024layer}
Guy Lorberbom, Itai Gat, Yossi Adi, Alexander Schwing, and Tamir Hazan.
\newblock Layer collaboration in the forward-forward algorithm.
\newblock In {\em Proceedings of the AAAI Conference on Artificial Intelligence}, volume~38, pages 14141--14148, 2024.

\bibitem{lecun1998gradient}
Yann LeCun, L{\'e}on Bottou, Yoshua Bengio, and Patrick Haffner.
\newblock Gradient-based learning applied to document recognition.
\newblock {\em Proceedings of the IEEE}, 86(11):2278--2324, 1998.

\bibitem{krizhevsky2009learning}
Alex Krizhevsky, Geoffrey Hinton, et~al.
\newblock Learning multiple layers of features from tiny images.
\newblock 2009.

\bibitem{deng2009imagenet}
Jia Deng, Wei Dong, Richard Socher, Li-Jia Li, Kai Li, and Li~Fei-Fei.
\newblock Imagenet: A large-scale hierarchical image database.
\newblock In {\em 2009 IEEE conference on computer vision and pattern recognition}, pages 248--255. Ieee, 2009.

\bibitem{yuan2021tokens}
Li~Yuan, Yunpeng Chen, Tao Wang, Weihao Yu, Yujun Shi, Zi-Hang Jiang, Francis~EH Tay, Jiashi Feng, and Shuicheng Yan.
\newblock Tokens-to-token vit: Training vision transformers from scratch on imagenet.
\newblock In {\em Proceedings of the IEEE/CVF international conference on computer vision}, pages 558--567, 2021.

\bibitem{dosovitskiy2020image}
Alexey Dosovitskiy, Lucas Beyer, Alexander Kolesnikov, Dirk Weissenborn, Xiaohua Zhai, Thomas Unterthiner, Mostafa Dehghani, Matthias Minderer, G~Heigold, S~Gelly, et~al.
\newblock An image is worth 16x16 words: Transformers for image recognition at scale.
\newblock In {\em International Conference on Learning Representations}, 2021.

\end{thebibliography}
% \bibliographystyle{iclr2026_conference}
%%%%%%%%%%%%%%%%%%%%%%%%%%%%%%%%%%%%%%%%%%%%%%%%%%%%%%%%%%%%

\newpage

\appendix 
\section{Appendix}
\renewcommand{\thesubsection}{A.\arabic{subsection}}
\renewcommand{\thefigure}{A\arabic{figure}}
\renewcommand{\thetable}{A\arabic{table}}

\vspace{0.5em}
\noindent \textbf{Appendix Contents}
\begin{itemize}
    \item \hyperref[sec:apdx_alg]{A.1 Algorithm and code}
    \item \hyperref[sec:apdx_res]{A.2 Additional Experimental Results}
    \item \hyperref[sec:apdx_proofs]{A.3 Proofs of Theorems}
    \item \hyperref[sec:apdx_exp]{A.4 Model Architecture and Experiment Details}
    % \item \hyperref[sec:apdx_ethic]{A.5 Ethic Statement}
\end{itemize}

\subsection{Algorithm and code}
\label{sec:apdx_alg}
\subsubsection{pseudo code}
\begin{algorithm}
\caption{Stochastic Layer-wise Learning}
\begin{algorithmic}[1]
    \Require Training batch data $(x, y)$, learning rate $\eta$, random projection matrix $R_l\sim \mathcal{N}$
    \Ensure Updated network weights $\theta$

    % \State \textbf{Forward Pass:}
    \For{each layer $l$ from $1$ to $L-1$}
        \State \textbf{Detach From above layer: } $h_{l-1} = h_{l-1}.detach()$ 
        % \State $h_{l-1} = h_{l-1}.detach()$ 
        % \State $z_l \gets W_l h_{l-1} + b_l$
        \State \textbf{Update activation: } $h_l \gets f(h_{l-1},\theta_l)$

        \State \textbf{Approximate $\partial h_l$:}
        \State \quad \quad Random Projection: $v_l =  dp(R_l)h_l$ or $v_l =  dp(R_l)[h_l,y]$
        \State \quad \quad Loss: $\mathcal{L}_l  \gets L_{Pred}(v_l, y) + L_{BC}(\mathrm{softmax}(v_l),\mathrm{softmax}(v_{l-1}))$
        \State \quad \quad Activation drift: $\varepsilon_l  \gets \frac{\partial \mathcal{L}_l}{\partial h_l}$
        
        % \State \textbf{Weight Update: } $\theta_l \gets \theta_l - \eta \cdot \frac{\partial \Delta h_l}{\partial \theta_l}$
        \State \textbf{Weight Update: } $\theta_l \gets \theta_l - \eta \cdot \frac{\partial h_l}{\partial \theta_l}\varepsilon_l$
        \State Clear unnecessary tensors 
        % - [optional: reduce memory, but increase the training time]
        % \State $\Delta\theta_l  \gets \frac{\partial \Delta h_l}{\partial \theta_l}$
        % \State $\theta_l \gets \theta_l - \eta \cdot \frac{\partial \Delta h_l}{\partial \theta_l}$
    
    \EndFor
    % \State Last layer updates:
    \State $h_{L-1} = h_{L-1}.detach()$
    \State $h_L \gets f(h_{L-1},\theta_L)$
    \State Loss: $\mathcal{L}_L  \gets L_{Pred}(h_L, y) + L_{fa}(h_L,v_{L-1})$
    \State $\theta_L \gets \theta_L - \eta \cdot \frac{\partial \mathcal{L}_L}{\partial \theta_L}$
\end{algorithmic}
\end{algorithm}
Activation drift is The first term (blue) captures the global contributions of activation $h_i$ to the global loss in Eq:2.

\subsubsection{Python code for Bhattacharyya coefficient}
\begin{lstlisting}[style=py,caption={Loss function},label={lst:helper}]
def L_BC_per(q: torch.Tensor, p: torch.Tensor,
                      reduction: str = "mean",
                      eps: float = 1e-12,
                      detach_p: bool = True) -> torch.Tensor:
    """
    q, p: [B, K] probabilities (nonnegative; rows ~ sum to 1).
    """
    if detach_p:
        p = p.detach()
    q = (q.clamp_min(eps) / q.clamp_min(eps).sum(dim=-1, keepdim=True))
    p = (p.clamp_min(eps) / p.clamp_min(eps).sum(dim=-1, keepdim=True))
    # log BC via log-sum-exp for stability
    log_bc = torch.logsumexp(0.5 * (q.log() + p.log()), dim=-1)
    loss_per = -log_bc
    return (loss_per.mean() if reduction == "mean"
            else loss_per.sum() if reduction == "sum"
            else loss_per)
\end{lstlisting}

\subsection{Additional Experimental Results}
\label{sec:apdx_res}
\subsubsection{Small CNN}

\begin{table}[H]
    \centering
    \begin{tabular}{l l c c}
        \toprule
        \textbf{Method} & \textbf{Model} & \textbf{CIFAR-10} & \textbf{CIFAR-100} \\ \hline
        \midrule
        BP     & CNN\cite{apolinario2024lls} &87.57 $\pm$ 0.13 & 62.25 $\pm$ 0.29 \\ \hline
        DFA    & CNN\cite{apolinario2024lls} &71.53 $\pm$ 0.38 & 44.93 $\pm$ 0.52\\
        PEPITA & CNN\cite{dellaferrera2022error} & 56.33 $\pm$ 1.35& 27.56 $\pm$ 0.60\\
        LLS    & CNN\cite{apolinario2024lls} &84.10 $\pm$ 0.27 & 55.32 $\pm$ 0.38\\ \hline
        SVP    & CNN & 87.48 $\pm$ 0.32 & 59.74 $\pm$ 0.27\\
        \bottomrule
    \end{tabular}
    \caption{Comparison of different methods on CIFAR-10 and CIFAR-100 using a 3-layer CNN. All methods are evaluated under the same network structure for a fair comparison.}
    \label{tab:small-cnn}
\end{table}

\subsubsection{Ablation study on CIFAR-100}
\begin{figure}[H]
	\centering
    \includegraphics[width=.7\textwidth]{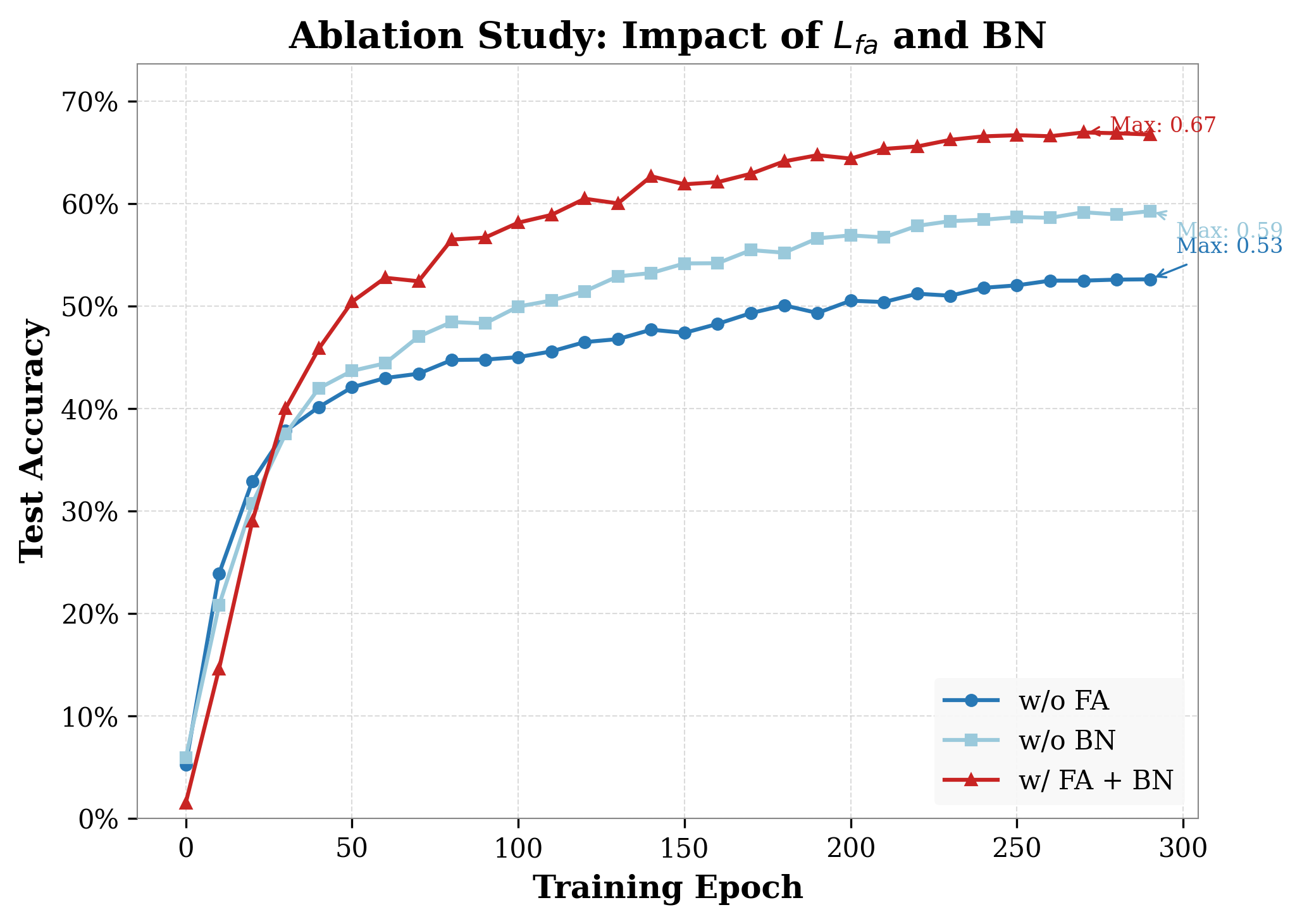}
	\caption{Ablation on CIFAR-100 showing that combining feature alignment loss ($\mathcal{L}_{\text{fa}}$) and batch normalization significantly improves test accuracy and convergence over variants without $\mathcal{L}_{\text{fa}}$ or BN. }
	\label{apdx_fig:c100abl}
    
\end{figure}

\begin{figure}[H]
	\centering
    \includegraphics[width=.7\textwidth]{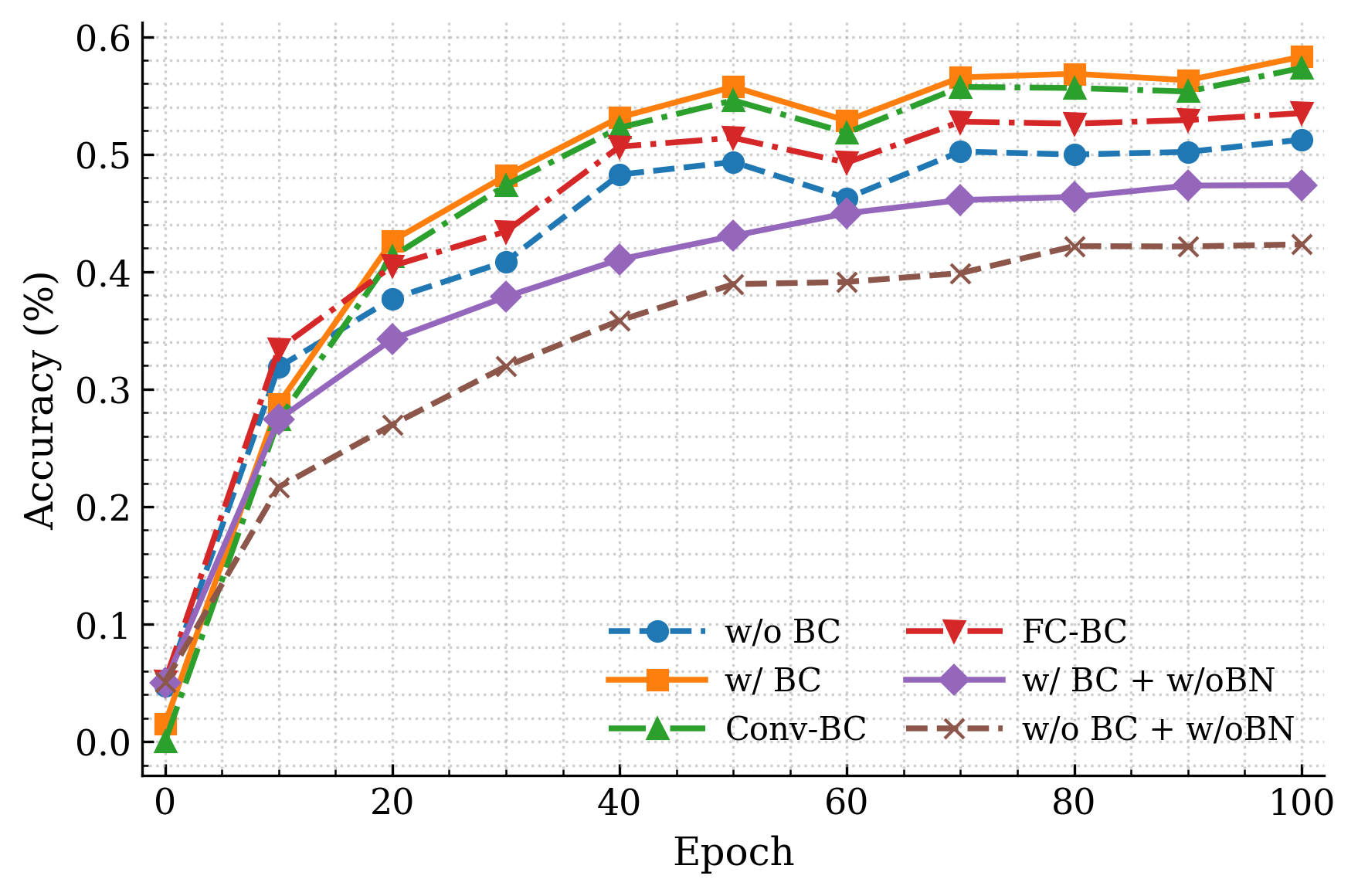}
    \vspace{-6pt}
	\caption{\textbf{ Ablations of the Bhattacharyya alignment loss $\mathcal{L}_{\mathrm{BC}}$.}
Top-1 accuracy versus epoch for the VGG model. 
\emph{w/o BC}: baseline. 
\emph{w/ BC}: $\mathcal{L}_{\mathrm{BC}}$ at all layers. 
\emph{Conv-BC} / \emph{FC-BC}: $\mathcal{L}_{\mathrm{BC}}$ applied only to convolutional or only to fully connected layers. 
\emph{w/ BC + w/o BN} and \emph{w/o BC + w/o BN}: BatchNorm removed to test interaction with $\mathcal{L}_{\mathrm{BC}}$. 
Adding $\mathcal{L}_{\mathrm{BC}}$ consistently accelerates learning and improves final accuracy; applying it on convolutional blocks (\emph{Conv-BC}) yields the strongest gains, while removing BatchNorm degrades performance that is partially recovered by $\mathcal{L}_{\mathrm{BC}}$.
}
	\label{apdx_fig:c100abl}
    
\end{figure}

\subsubsection{Layer-wise loss and Accuracy}
\begin{figure}[H]
	\centering
    \includegraphics[width=.8\textwidth]{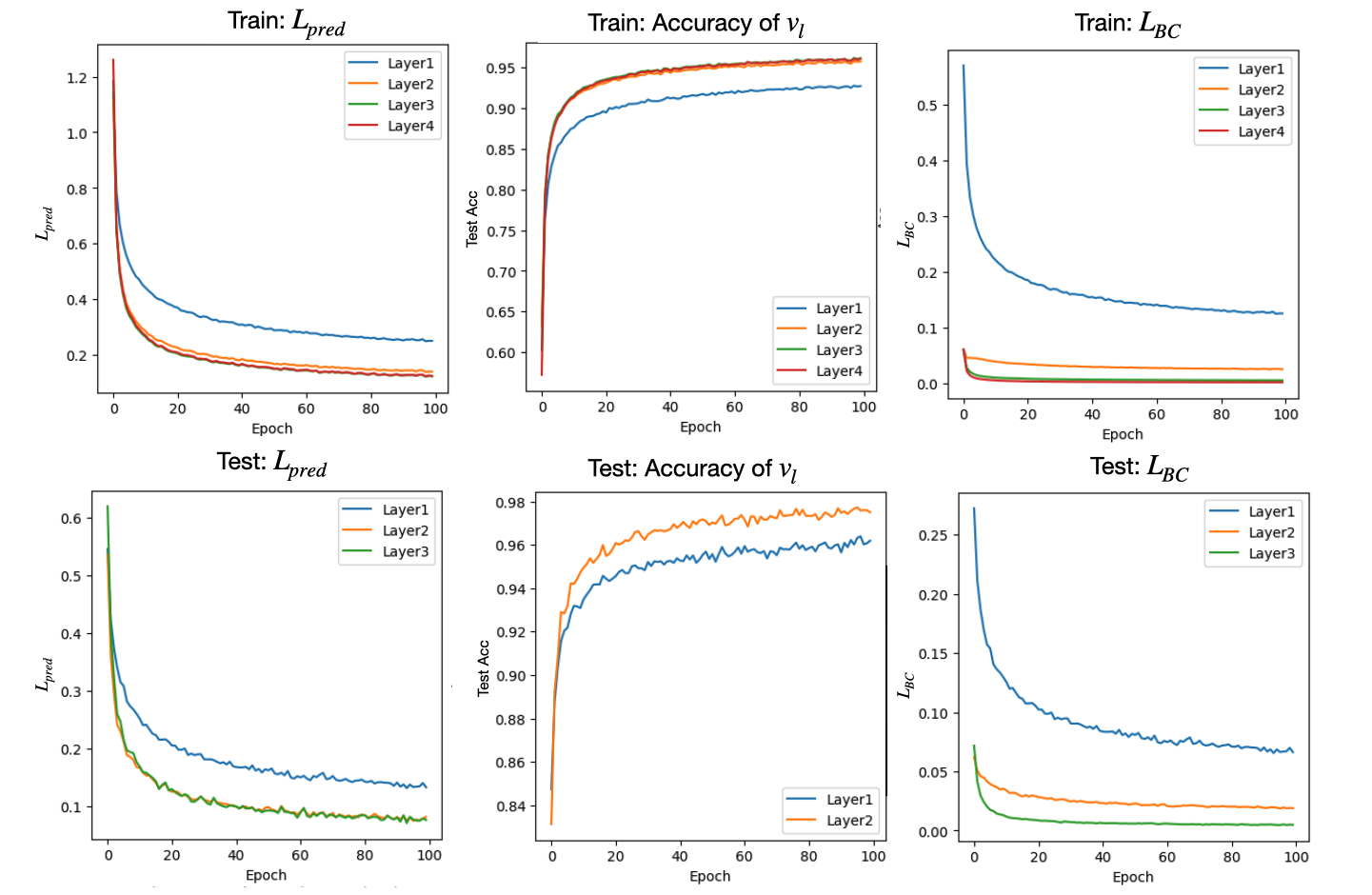}
	\caption{\textbf{Layer-wise training dynamics under SLL.}
    Top: Learning curves for training dataset; bottom: earning curves for test dataset. 
    Left: prediction loss $\mathcal{L}_{\text{pred}}$ on projected codes $v_{\ell}=R_{\ell}h_{\ell}$. 
    Middle: classification accuracy from the head on $v_{\ell}$. 
    Right: Bhattacharyya alignment loss $\mathcal{L}_{\mathrm{BC}}$ between induced posteriors $(q_{\ell},p_{\ell})$. 
    Curves are shown per layer; deeper layers achieve lower $\mathcal{L}_{\text{pred}}$ and $\mathcal{L}_{\mathrm{BC}}$ and higher accuracy, indicating progressive local learning and strengthened inter-layer consistency without cross-layer backpropagation. }
	\label{apdx_fig:m10_loss_acc}
    
\end{figure}

% \subsubsection{MNIST feature visualization}

% \begin{figure}[H]
% 	\centering
% 	% \includegraphics[width=.95\textwidth]{figures/mlp_c10_tsne.png}
%     \includegraphics[width=.95\textwidth]{figures/tsne-m10.png}
% 	\caption{t-SNE visualization of MNIST representations. Each color indicates a digit class. \textbf{Top}: Layer-wise activations  $h_i$ from a network trained with BP. \textbf{Middle}: Activations from a network trained with SVP. \textbf{Bottom}: Corresponding random projections $v_i = R_i h_i$ from SVP. SVP leads to more compact and class-separable features, with random projections preserving and enhancing this structure. }
% 	\label{apdx_fig:tsne-mlp-m10}
    
% \end{figure}

\subsection{Proof of Theorems}
\label{sec:apdx_proofs}
\subsubsection{Theorem 1: Layer-wise ELBO Provides a Valid Variational Bound}
% \paragraph{Theorem 1: Layer-wise ELBO Provides a Valid Variational Bound} 
\textit{Let $\mathcal{L}_{NN}$ be the global Evidence Lower Bound (ELBO) of the network:}
\begin{equation}
    \mathcal{E}_{NN} = \mathbb{E}_q [\log p(y \mid h_L)] - D_{\text{KL}}(q(\mathcal{H}) \| p(\mathcal{H})).
\end{equation}
\textit{Then, the sum of layer-wise ELBOs in SVP provides a valid lower bound:}
\begin{equation}
    \frac{1}{L}\sum_{i=1}^{L} \mathcal{E}_i \leq \mathcal{E}_{NN}.
\end{equation}

\textit{Proof.} We start with the marginal likelihood:
\begin{equation}
\log p(y | x) = \log \int p(y, h_1, h_2, ..., h_L | x) \, dh_1 \dots dh_L.
\end{equation} Here, we define $\mathcal{H} =\mathcal{H}_L = \{h_1, h_2, ..., h_L\}$ as the set of activations. Introducing the variational approximation $q(\mathcal{H} | x, y)$, we have:
\begin{equation}
\log p(y | x) = \log \int \frac{p(y, \mathcal{H} | x)}{q(\mathcal{H} | x, y)} q(\mathcal{H} | x, y) \, dh_1 \dots dh_L.
\end{equation}

Applying Jensen's inequality (since logarithm is a concave function):
\begin{equation}
\log p(y | x) \geq \int \log \frac{p(y, \mathcal{H} | x)}{q(\mathcal{H} | x, y)} q(\mathcal{H} | x, y) \, dh_1 \dots dh_L = \mathbb{E}_{q(\mathcal{H}|x,y)} \left[ \log \frac{p(y, \mathcal{H} | x)}{q(\mathcal{H} | x, y)} \right].
\end{equation}

Using the joint probability factorization:
\begin{equation}
p(y, \mathcal{H} | x) =  \prod_{i=1}^{L+1} p(h_i | h_{i-1}, x),
\end{equation}
where $h_0 = x$ and $h_{L+1} = y$ by convention.

Thus, we obtain the global ELBO:
\begin{equation}
\mathcal{E}_{NN} = \mathbb{E}_{q(\mathcal{H}|x,y)} \left[ \log p(y | h_L, x) + \sum_{i=1}^{L} \log p(h_i | h_{i-1}, x) - \log q(\mathcal{H} | x, y) \right].
\end{equation}

Using the variational factorization assumption in SLL: 
\begin{equation}
q(\mathcal{H} | x, y) = \prod_{i=1}^{L} q(h_i | h_{i-1}, x, y),
\tag{Assumption 2}
\end{equation}
where again $h_0 = x$.

From the \textit{Assumption 1,2},  the global ELBO can be rewritten as:
\begin{align}
& \mathcal{E}_{NN} = \mathbb{E}_{q(\mathcal{H}|x,y)} \left[ \log p(y | h_L, x) + \sum_{i=1}^{L} \log p(h_i | h_{i-1}, x) - \sum_{i=1}^{L} \log q(h_i | h_{i-1}, x, y) \right] \\
& = \mathbb{E}_{q(\mathcal{H}|x,y)} \left[ \log p(y | h_L, x) \right] + \sum_{i=1}^{L} \mathbb{E}_{q(\mathcal{H}|x,y)} \left[ \log p(h_i | h_{i-1}, x) - \log q(h_i | h_{i-1}, x, y) \right].
\end{align}
We can rewrite this in terms of KL divergence:
\begin{equation}
\mathcal{E}_{NN} = \mathbb{E}_{q(\mathcal{H}|x,y)} \left[ \log p(y | h_L, x) \right] - \sum_{i=1}^{L} \mathbb{E}_{q(\mathcal{H}|x,y)} \left[ \log q(h_i | h_{i-1}, x, y) - \log p(h_i | h_{i-1}, x) \right].
\end{equation}

where the expectation of the KL divergence terms can be rewritten as:
\begin{align}
&\mathbb{E}_{q(\mathcal{H}|x,y)} \left[ \log q(h_i | h_{i-1}, x, y) - \log p(h_i | h_{i-1}, x) \right]  \\
= & \mathbb{E}_{q(\mathcal{H}_i|x,y)} \left[ D_{\text{KL}}(q(h_i | h_{i-1}, x, y) \| p(h_i | h_{i-1}, x)) \right].
\end{align}

Therefore, the global ELBO becomes:
\begin{equation}
\mathcal{E}_{NN} = \mathbb{E}_{q(\mathcal{H}|x,y)} \left[ \log p(y | h_L, x) \right] - \sum_{i=1}^{L} \mathbb{E}_{q(\mathcal{H}_i|x,y)} \left[ D_{\text{KL}}(q(h_i | h_{i-1}, x, y) \| p(h_i | h_{i-1}, x)) \right].
\end{equation}

Now, let's define the layer-wise ELBO for each layer $i$:
\begin{equation}
\mathcal{E}_i = \mathbb{E}_{q(\mathcal{H}_i|x,y)} \left[ \log p(y | h_i, x) \right] - \mathbb{E}_{q(\mathcal{H}_i|x,y)} \left[ D_{\text{KL}}(q(h_i | h_{i-1}, x, y) \| p(h_i | h_{i-1}, x)) \right].
\end{equation}

Summing the layer-wise ELBOs:
\begin{equation}
\sum_{i=1}^{L} \mathcal{E}_i = \sum_{i=1}^{L} \mathbb{E}_{q(\mathcal{H}_i|x,y)} \left[ \log p(y | h_i, x) \right] - \sum_{i=1}^{L} \mathbb{E}_{q(\mathcal{H}_i|x,y)} \left[ D_{\text{KL}}(q(h_i | h_{i-1}, x, y) \| p(h_i | h_{i-1}, x)) \right].
\end{equation}

\paragraph{Assumption 3:  Monotone predictive gain under consistent measure.}  This assumption is well-founded in established theoretical results showing that neural network expressivity grows exponentially with depth, with deeper representations providing exponentially more representational capacity than shallow ones~\cite{telgarsky2016benefits, eldan2016power}. Then we assume:
For all $i < L$:
\begin{equation}
\mathbb{E}_{q(h_i|x,y)} [\log p(y | h_i, x)] \leq \mathbb{E}_{q(h_L|x,y)} [\log p(y | h_L, x)]
\tag{Assumption 3}
\end{equation}

\textbf{Assumption 4:  KL budget constraint.} This constraint ensures that the accumulated KL regularization cost across all layers does not exceed the total predictive improvement gained from using the full network depth, preventing the variational bound from becoming arbitrarily loose due to excessive regularization. So, we assume:
\begin{equation}
\begin{split}
&\frac{L-1}{L} \sum_{i=1}^{L} \mathbb{E}_{q(h_{i-1}|x,y)} \left[ D_{\text{KL}}\big(q(h_i | h_{i-1}, x, y) \| p(h_i | h_{i-1}, x)\big) \right] \nonumber \\
&\quad \leq \frac{1}{L} \sum_{i=1}^{L} \left( \mathbb{E}_{q(h_L|x,y)} [\log p(y | h_L, x)] - \mathbb{E}_{q(h_i|x,y)} [\log p(y | h_i, x)] \right)
\end{split}
\tag{Assumption 4}
\end{equation}

\paragraph{Proof.}
\textbf{Step 1}: Using the Markov factorization, the global ELBO becomes:
\begin{align}
\mathcal{E}_{NN} &= \mathbb{E}_{q(h_L|x,y)} [\log p(y | h_L, x)] \nonumber \\
&\quad - \sum_{i=1}^{L} \mathbb{E}_{q(h_{i-1}|x,y)} \left[ D_{\text{KL}}\big(q(h_i | h_{i-1}, x, y) \| p(h_i | h_{i-1}, x)\big) \right]
\end{align}

where $q(h_L|x,y)$ is the marginal of the joint distribution $q(\mathcal{H}|x,y)$ under the Markov factorization.

\textbf{Step 2}: Define 
\begin{align}
A_i &:= \mathbb{E}_{q(h_i|x,y)} [\log p(y | h_i, x)] \\
K_i &:= \mathbb{E}_{q(h_{i-1}|x,y)} \left[ D_{\text{KL}}\big(q(h_i | h_{i-1}, x, y) \| p(h_i | h_{i-1}, x)\big) \right] \geq 0
\end{align}

Then:
\begin{align}
\mathcal{E}_i &= A_i - K_i \\
\mathcal{E}_{NN} &= A_L - \sum_{i=1}^L K_i
\end{align}

\textbf{Step 3}: Compute the difference:
\begin{align}
\frac{1}{L}\sum_{i=1}^{L} \mathcal{E}_i - \mathcal{E}_{NN} &= \frac{1}{L}\sum_{i=1}^{L} (A_i - K_i) - \left(A_L - \sum_{i=1}^L K_i\right) \\
&= \frac{1}{L}\sum_{i=1}^{L} A_i - A_L - \frac{1}{L}\sum_{i=1}^{L} K_i + \sum_{i=1}^L K_i \\
&= \frac{1}{L}\sum_{i=1}^{L} A_i - A_L + \frac{L-1}{L}\sum_{i=1}^{L} K_i \\
&= -\frac{1}{L}\sum_{i=1}^{L}(A_L - A_i) + \frac{L-1}{L}\sum_{i=1}^{L} K_i
\end{align}

\textbf{Step 4}: Apply the assumptions:
\begin{itemize}
\item By \textbf{Assumption 3}: Each $A_L - A_i \geq 0$, so $-\frac{1}{L}\sum_{i=1}^{L}(A_L - A_i) \leq 0$
\item By \textbf{Assumption 4}: $\frac{L-1}{L}\sum_{i=1}^{L} K_i \leq \frac{1}{L}\sum_{i=1}^{L}(A_L - A_i)$.  Justification: On average across layers, the extra coding cost---quantified by the KL between $q$ and $p$, i.e., $\tfrac{L-1}{L}\sum_{i=1}^{L} K_i$---of shaping the latents must be no larger than the predictive gain---the increase in expected log-likelihood, i.e., $\tfrac{1}{L}\sum_{i=1}^{L} (A_L - A_i)$. In short, depth’s benefit pays for its inference complexity.

\end{itemize}

\textbf{Step 5}: Conclude:
\begin{equation}
\frac{1}{L}\sum_{i=1}^{L} \mathcal{E}_i - \mathcal{E}_{NN} = -\frac{1}{L}\sum_{i=1}^{L}(A_L - A_i) + \frac{L-1}{L}\sum_{i=1}^{L} K_i \leq 0
\end{equation}

Therefore: $\frac{1}{L}\sum_{i=1}^{L} \mathcal{E}_i \leq \mathcal{E}_{NN}$

This completes the proof that the mean of layer-wise ELBOs provides a valid lower bound for the global ELBO.

\subsubsection{Theorem 2: MI preservation of Random Projection} 
\textit{Let $h_i = f_i(h_{i-1}) \in \mathbb{R}^d$ be the deterministic activation at layer $i$, and define the projected representation $v_i = R_i h_i$ using the fixed random matrix $R_i \sim \mathcal{N}(0, 1/d')$. Assume $h_i$ has bounded second moments. Let $n$ denote the effective number of distinct activation patterns in the support of the distribution of $h_i$." Then, for any $\varepsilon \in (0,1)$, if $d' = O(\varepsilon^{-2} \log n)$, the mutual information satisfies, with high probability over $R_i$,
\begin{equation}
I(x; v_i) \geq I(x; h_i) - \delta(\varepsilon),
\end{equation}
where $\delta(\varepsilon) \to 0$ as $\varepsilon \to 0$, and $x$ denotes the network input.}

\textit{Proof.} For any random variables $X$ and $Y$, the mutual information is defined as:
\begin{equation}
I(X; Y) = H(Y) - H(Y|X)
\end{equation}

Since $h_i$ is a deterministic function of $x$, we have $H(h_i|x) = 0$, thus:
\begin{equation}
I(x; h_i) = H(h_i)
\end{equation}

For the projected representation, given $x$, $v_i|x$ follows a Gaussian distribution with entropy:
\begin{equation}
H(v_i|x) = \frac{d'}{2}\log(2\pi e \frac{|h_i(x)|^2}{d'})
\end{equation}

By the Johnson-Lindenstrauss lemma~\cite{johnson1984extensions}, for any two points $h_i(x_1)$ and $h_i(x_2)$, with high probability:
\begin{equation}
(1-\varepsilon)\|h_i(x_1) - h_i(x_2)\|^2 \leq \|v_i(x_1) - v_i(x_2)\|^2 \leq (1+\varepsilon)\|h_i(x_1) - h_i(x_2)\|^2
\end{equation}
when $d' = O(\varepsilon^{-2}\log n)$.

For Gaussian distributions with covariance $\Sigma = \frac{1}{d'}I_{d'}$, the KL divergence is:
\begin{equation}
D_{KL}(p(v_i|x_1) | p(v_i|x_2)) = \frac{d'}{2}\|R_i h_i(x_1) - R_i h_i(x_2)\|^2
\end{equation}

The preservation of distances directly implies preservation of distinguishability between different inputs. By discretizing the spaces of $h_i$ and $v_i$ and applying quantization theory, we can derive:
\begin{equation}
I(x; v_i) \geq (1-\varepsilon')I(x; h_i) - C\varepsilon'\log(1/\varepsilon')
\end{equation}
where $\varepsilon' = O(\varepsilon)$ and $C$ is a constant depending on the bounded second moments of $h_i$.

Define $\delta(\varepsilon) = \varepsilon'I(x; h_i) + C\varepsilon'\log(1/\varepsilon')$. Since $I(x; h_i)$ is bounded by our assumptions, $\delta(\varepsilon) \to 0$ as $\varepsilon \to 0$. Therefore, with high probability over the choice of $R_i$:
\begin{equation}
I(x; v_i) \geq I(x; h_i) - \delta(\varepsilon)
\end{equation}
where $\delta(\varepsilon) \to 0$ as $\varepsilon \to 0$.

\subsection{Model Architecture and Experiment Details}
\label{sec:apdx_exp}
\subsubsection{Computer Resources}
All experiments were conducted on a single NVIDIA A100 GPU with 40GB of memory. No multi-GPU or distributed training was used.

\subsubsection{Datasets}
In this paper, we evaluate SVP on a range of benchmark datasets.
\begin{itemize}
    \item MNIST is a handwritten digit image dataset over 10 classes including 60,000 images for training and 10,000 images for testing. Each image is a 28 × 28 gray-scale image.
    \item Fashion MNIST contains fashion items images such as clothing and shoes. It consists of a training set of 60,000 grayscale images and a test set of 10,000 images. Each image has a 28 × 28 size and is categorized into 10
    classes. 
    \item CIFAR-10 consists of 32 × 32 RGB images for object recognition with 50,000 images for training and 10,000 images for testing. It has 10 classes,  with 5{,}000 training and 1{,}000 testing images per class.
    \item CIFAR-100 comprises a total of 60,000 32 × 32 RGB images distributed across 100 classes. Within each class, 500 images are allocated for training, while 100 images are for testing. 
    \item Tiny-ImageNet is a downsampled subset of ImageNet to a size of 64 × 64. This dataset consists of 200 classes and each class contains 500 images for training and 100 images for testing.
    \item ImageNet-1Kis a large-scale image classification dataset containing over 1.28 million training images and 50{,}000 validation images across 1{,}000 classes. Images are typically resized to 224\,×\,224 pixels for training.
\end{itemize}

\subsubsection{MLP}
\paragraph{Architecture}
For MNIST, we use a 2-layer MLP with 800 neurons per layer and ReLU activations, followed by dropout.
For CIFAR-10 and CIFAR-100, we adopt a 3-layer MLP with 1000 neurons per layer, also using ReLU activations and dropout regularization.

\paragraph{Experimental Details.}
During training, we apply random horizontal flipping and normalization as standard data augmentation. Models are optimized using the Adamax optimizer with a learning rate of 0.001, trained for 100 epochs across all datasets.

\subsubsection{CNN}
The architecture and training configurations used for CNN experiments are summarized in Table~\ref{tab:cnn_training_details}.
\begin{table}[H]
\centering
\begin{tabular}{|c|p{6cm}|p{4cm}|}
\hline
\textbf{Dataset} & \textbf{Network Architecture} & \textbf{Training Details} \\
\hline
\begin{tabular}{c}
FMNIST \\
SIZE: 28$\times$28\\
CLASS: 10 \\
\end{tabular}& 
Conv64k3 - MaxPool2 - Conv128k3 - MaxPool2 - Conv256k3 - Conv256k3 - MaxPool2 - Conv512k3 - MaxPool2 - FC10 & 
Data Aug: Normalize \newline
Optimizer: Adam \newline
Learning Rate: 0.003 \newline
Batch Size: 128 \newline
Epochs: 100 \\
\hline
\begin{tabular}{c}
CIFAR10/100 \\
SIZE: 32$\times$32\\
CLASS: 100 \\
\end{tabular}
& 
Conv256k3 - MaxPool2 - Conv512k3 - MaxPool2 - Conv512k3 - Conv1024k3 - MaxPool2 - Conv1024k3 - MaxPool2 - Conv1024k3 - Conv1024k3 - FC2048 - FC100 & 
Data Aug: Crop, Flip, AutoAugment, Normalize \newline
Optimizer: Adam \newline
Learning Rate: 0.003 \newline
Batch Size: 128 \newline
Epochs: 500 \\
\hline
\begin{tabular}{c}
Imagenette \\
SIZE: 224$\times$224\\
CLASS: 10 \\
\end{tabular}
& 
Conv64k3 - MaxPool2 - Conv128k3 - MaxPool2 - Conv256k3 - Conv256k3 - MaxPool2 - Conv512k3 - MaxPool2 - Conv512k3 - Conv1024k3 - FC10 & 
Data Aug: Crop, Flip, Normalize \newline
Optimizer: Adam \newline
Learning Rate: 0.001 \newline
Batch Size: 128 \newline
Epochs: 500 \\
\hline

\begin{tabular}{c}
Tiny-Imagenet \\
SIZE: 64$\times$64\\
CLASS: 200 \\
\end{tabular}
& 
Conv64k3 - MaxPool2 - Conv128k3 - MaxPool2 - Conv256k3 - Conv256k3 - MaxPool2 - Conv512k3 - MaxPool2 - Conv512k3 - Conv1024k3 - FC4096 - FC200& 
Data Aug: Crop, Flip, Normalize \newline
Optimizer: Adam \newline
Learning Rate: 0.001 \newline
Batch Size: 128 \newline
Epochs: 500 \\
\hline
\end{tabular}
\caption{Network architectures and training configurations for CNNs.}
\label{tab:cnn_training_details}
\end{table}

\subsubsection{VIT}
Details of the ViT architecture and experimental setup are provided in Table~\ref{tab:vit_training_details}.

\begin{table}[H]
\centering
\begin{tabular}{|c|p{5cm}|p{4cm}|}

\hline
\begin{tabular}{c}
CIFAR10/100 \\
SIZE: 32$\times$32\\
CLASS: 100 \\
\end{tabular}
& 
blocks=7,\newline heads=12,\newline mlp-ratio=2,\newline embedding=384 & 
Data Aug: Crop, Flip, AutoAugment, Normalize \newline
Optimizer: AdamW \newline
Learning Rate: 0.003 \newline
Batch Size: 128 \newline
Epochs: 500 \\
\hline
\begin{tabular}{c}
Imagenette \\
SIZE: 224$\times$224\\
CLASS: 10 \\
\end{tabular}
& 
blocks=7, \newline heads=12,\newline mlp-ratio=2,\newline embedding=384 & 
Data Aug: Crop, Flip, Normalize \newline
Optimizer: AdamW \newline
Learning Rate: 0.001 \newline
Batch Size: 128 \newline
Epochs: 500 \\
\hline

\begin{tabular}{c}
Imagenet \\
SIZE: 224$\times$224\\
CLASS: 1000 \\
\end{tabular}
& 
blocks=12,\newline heads=6,\newline mlp-ratio=3,\newline embedding=384& 
Data Aug: Crop, Flip, Normalize \newline
Optimizer: AdamW \newline
Learning Rate: 0.0005 \newline
Batch Size: 256 \newline
Epochs: 500 \\
\hline
\end{tabular}
\caption{Network architectures and training configurations for ViTs.}
\label{tab:vit_training_details}
\end{table}

% \paragraph{Architecture}

% \paragraph{Experiments details}

\end{document}